\documentclass[10pt,twocolumn,letterpaper]{article}

\usepackage[pagenumbers]{cvpr} 

\usepackage{graphicx}
\usepackage{amsmath}
\usepackage{multirow}
\usepackage{booktabs}
\usepackage{amssymb}

\usepackage[pagebackref,breaklinks,colorlinks]{hyperref}

\usepackage[capitalize]{cleveref}
\crefname{section}{Sec.}{Secs.}
\Crefname{section}{Section}{Sections}
\Crefname{table}{Table}{Tables}
\crefname{table}{Tab.}{Tabs.}

\def\cvprPaperID{****} 
\def\confName{CVPR}
\def\confYear{2022}

\begin{document}

\title{VLDeformer: Vision-Language Decomposed Transformer for Fast Cross-Modal Retrieval}

\author{Lisai Zhang$^1$, Hongfa Wu$^1$, Qingcai Chen$^{*1}$, Yimeng Deng$^{*2}$, Joanna Siebert$^1$, Zhonghua Li$^2$ \and Yunpeng Han$^1$, Dejiang Kong$^2$,  Zhao Cao$^2$ \\
$^1$Harbin Institute of Technology, Shezhen, $^2$Huawei Technologies Co., Ltd \\
{\tt\small \{LisaiZhang,hongfawu\}@foxmail.com, qingcai.chen@hit.edu.cn} \\
{\tt\small	\{dengyimeng,lizhonghua3,kongdejiang,caozhao1\}huawei.com}
}
\maketitle

\begin{abstract}
Cross-model retrieval has emerged as one of the most important upgrades for text-only search engines (SE). Recently, with powerful representation for pairwise text-image inputs via early interaction, the accuracy of vision-language (VL) transformers has outperformed existing methods for text-image retrieval. However, when the same paradigm is used for inference, the efficiency of the VL transformers is still too low to be applied in a real cross-modal SE. 
Inspired by the mechanism of human learning and using cross-modal knowledge, 
this paper presents a novel Vision-Language Decomposed Transformer (VLDeformer), which greatly increases the efficiency of VL transformers while maintaining their outstanding accuracy. 
By the proposed method, the cross-model retrieval is separated into two stages: the VL transformer learning stage, and the VL decomposition stage. The latter stage plays the role of single modal indexing, which is to some extent like the term indexing of a text SE.
The model learns cross-modal knowledge from early-interaction pre-training and is then decomposed into an individual encoder. 
The decomposition requires only small target datasets for supervision and achieves both $1000+$ times acceleration and less than $0.6$\% average recall drop.
VLDeformer also outperforms state-of-the-art visual-semantic embedding methods on COCO and Flickr30k.~\footnote{Our code will be publicly available upon acceptance.}
\end{abstract}


\section{Introduction}
Cross-modal retrieval is important for modern applications such as search engines, social media, and e-commerce, which involves searching for instances semantically similar to the query from another modal. Just like text search engines, cross-model retrieval requires not only high accuracy but also fast retrieval speed. 

Vision-language transformers  (VL transformers) ~\cite{vilbert,unicoder,imagebert,oscar} have been known for their superior accuracy in cross-modal retrieval. These models learn cross-modal matching relationships with a BERT network~\cite{bert} where the features are fused with early interaction dataflow. 
Similar to the process of human learning cross-modal alignment, most VL transformers compare a pair of text-images each time and produce one joint representation during pre-training. These models significantly outperform conventional visual-semantic embedding methods and prove the effectiveness of pre-training and early interaction. 

\begin{figure}
\centering
\small
\setlength{\tabcolsep}{0.1em}
\begin{tabular}{cc}
	\multicolumn{2}{c}{\includegraphics[width=\linewidth]{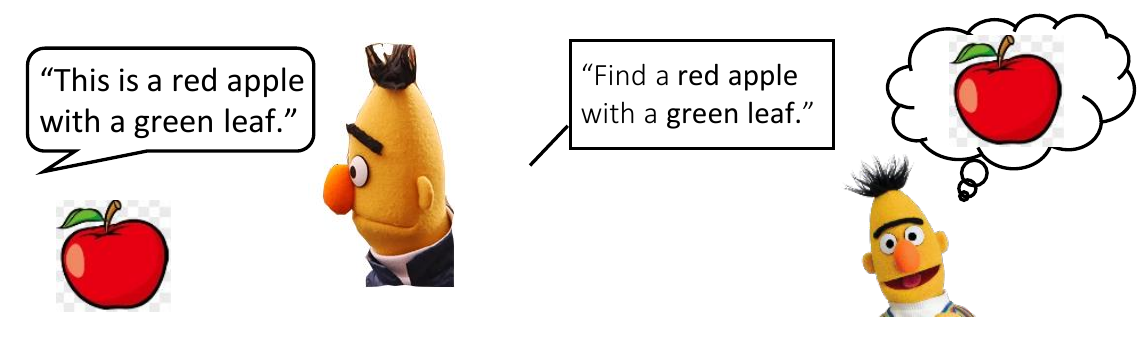}} \\
	\footnotesize Learning from pairwise input &
	\footnotesize Retrieval with individual input
\end{tabular}
\caption{Illustration of the pre-training and retrieval paradigm for VLDeformer. }
\label{fig.mode}
\vspace{-5pt}
\end{figure}

However, the VL transformers still keep the paradigm for inference, whose computing costs are too high to be applied in real applications. 
Because of the early-interaction dataflow, VL transformers have to compute a joint representation for every text-image combination for matching. The paradigm results in $O(n^2)$ inference times on text-image data. In practice, finding the most similar pairs from $1k$ text-image records requires $500k$ inferences, which costs about $30$ minutes on a modern V100 machine. 

Visual-semantic embeddings models~\cite{pfan_13,pfan_6,pfan_21,pfan_2,pfan++,sigir2021} are another kind of mainstream cross-model retrieval methods. They use two-branch encoders for single image and text input. In these models, late-interaction dataflow is used to learn the cross-modal alignment, which contributes to much faster retrieval efficiency than VL transformers. During retrieval, the embeddings could be cached and reused for other comparisons, which results in $O(n)$ inference times. The main weakness of these models is their relatively low accuracy.
Recent works~\cite{lightningdot,ALIGN} attempt to address the trade-off between accuracy and speed by pre-training. However, as shown in Table~\ref{tab.mode}, these models are still not as effective as VL transformers while using similar or even larger pre-training data. Until now, two-branch encoders are reported~\cite{ALIGN} to outperform VL transformers only using hundreds of times larger pre-training data. 
Therefore, VL transformers are still the most effective method for cross-modal retrieval but have shortcomings in retrieval speed.

The early-interaction pre-training of VL transformers is similar to humans learning the cross-modal alignment by observing pairwise language and vision inputs.
However, humans are able to handle individual text and image information separately after learning, as in Fig.~\ref{fig.mode}, which is a faster paradigm for retrieval. 
From this point of view, existing VL transformers only achieve the learning stage.
To inspect the feasibility of converting a VL transformer into an individual encoder for text and image, we analyze the dataflow of VinVL~\cite{vinvl} in a pilot experiment (Sec.~\ref{sec.pilot}). Surprisingly, we find an interesting phenomenon: most cross-modal interactions do not happen point-to-point, but through some ``routing nodes'', as illustrated in Fig.~\ref{fig.route}.  Moreover, the special tokens $\mathtt{[CLS]}$ and $\mathtt{[SEP]}$ are routing nodes most of the time. Since these special tokens do not belong to any specific modalities, it is possible to divide the dataflow by modality from them without breaking the learned dataflow. 
Following these findings, we prove that the pre-trained VL transformer could be decomposed into an individual encoder while maintaining most of the accuracy with only a small target dataset as supervision. 
We also verify in the experiment that the ``routing nodes'' are successfully kept after decomposition. 

Combined with the pre-training stage, the proposed Vision-Language Decomposed Transformer (VLDeformer) presents a new paradigm that pre-trains a VL transformer with early-interaction dataflow and then decomposes it into an individual encoder. Using this paradigm, we can build a VL transformer of not only high accuracy but also fast speed.
On the public COCO and Flickr30k benchmarks, VLDeformer achieves both $1000+$ times acceleration and less than $0.6$\% average recall drop for VL transformer.  VLDeformer also outperforms state-of-the-art visual-semantic embedding methods while using similar or even smaller pre-training data. 

\section{Related Work}

\begin{table}
	\centering
	\small
	\setlength{\tabcolsep}{3pt}
	\begin{tabular}{@{}lccccc@{}}
		\toprule
		Type & Model & T2I & I2T & Pre-train data & Time (s) \\ \midrule
		(a) & VinVL\cite{vinvl} & 74.6 & 58.1 & 8.8M & 32537.5 \\ \midrule
		(b) & LightningDOT\cite{lightningdot} & 70.0 & 54.0 & 9.5M & 7.4 \\
		(b) & ALIGN-small\cite{ALIGN} & 52.0 & 39.2 & 180M & - \\
		\bottomrule
	\end{tabular}
	\caption{R@1 retrieval score, pre-training data scale, and inference time of the state-of-the-art VL Transformer (a) and Visual-Semantic Embedding (b) models on COCO 5k test. }
	\label{tab.mode}
	\vspace{-10pt}
\end{table}

\textbf{VL Transformers} Pre-trained VL transformers have shown impressive performance for many multimodal tasks. These models learn the cross-modal interaction with an early-interaction dataflow, where the text and image features in each layer are fused with the attention mechanism. For example, VilBERT~\cite{vilbert} achieves early-interaction with two-branch transformer networks connected by co-attention. In the model, the outputs of the image and text of each layer are fused by a third transformer. The following models~\cite{visualbert,unicoder,uniter,vlbert,oscar,imagebert} use single-stream architecture where the features are fused with fully-connected attention mechanisms. These models achieve improvement on cross-modal retrieval tasks with different self-supervised tasks.
In common practice, the pre-trained VL transformers still keep the early-interaction dataflow while inferencing on downstream tasks, which has high computation costs in large-scale cross-modal retrieval tasks.

\textbf{Visual-Semantic Embedding Models}
Text-image retrieval methods~\cite{pfan_4,pfan_17,pfan_19} usually learn embeddings for the image and text with a two-branch network. Generally, the network includes a convolutional neural network as the image encoder and a sequence model as the text encoder. 
Researchers have found that the fine-grained relations between the visual objects and text tokens are essential for improving the embedding quality. For example, Lee et al.,\cite{pfan_2} calculates attention between detected object features and word embedding from both visual and text view, respectively. The following work SCAN~\cite{SCAN} explores early-interaction structure by stacked cross-attention and gets significant improvements. However, the early-interaction dataflow decreases its retrieval speed. During inference, the late-interaction dataflow enables extracting representations offline to achieve fast online retrieval. Therefore, most of the following visual-semantic embedding methods still use late-interaction dataflow; e.g, Qu et al.~\cite{CAAN} and Zhang et al.~\cite{sigir2021} develop the interaction from global and local views.

\textbf{Pre-trained Visual-Semantic Embedding Models}
Pre-training transformers typically use several millions of text-image pairs. 
Recently, some researchers have explored pre-training visual-semantic embedding methods with larger text-image pairs. For example, CLIP~\cite{CLIP} trains a two-branch network with contrastive learning on 400 million image text pairs and achieves competitive results on various downstream tasks. ALIGN~\cite{ALIGN} expands the pre-training data even further to a larger and noisier 1,800 million scale. The model consists of an EfficientNet~\cite{efficientnet} with global pooling as the image encoder and a BERT as the text encoder and achieves higher performance than the pre-trained VL transformers after fine-tuning. Although these pre-trained visual-semantic embedding models achieve advanced performance, their corpora are hundreds of times larger than VL transformers.
Inspired by the pre-training of VL transformers, LightingDOT~\cite{lightningdot} tries to train two-branch transformers with a data scale closer to the data scale used by the VL transformers. The network uses a late-interaction dataflow to accelerate the inference time.  However, there is still a performance gap from the VL transformers, which has to be made up by collecting top-$M$ samples and using a VL transformer to select the final top-$K$ ($M > K$) results.

In contrast with the above methods, we regard the early-interaction pre-training as the first stage to learning cross-modal knowledge, and the model could be decomposed as an individual encoder for fast inference. Following this process, we build a transformer that achieves both state-of-the-art accuracy and fast speed at the same time.

\section{Pilot Analysis of VL Transformer}
\label{sec.pilot}
\begin{figure}
	\centering
	\begin{tabular}{c}
		\includegraphics[width=\linewidth]{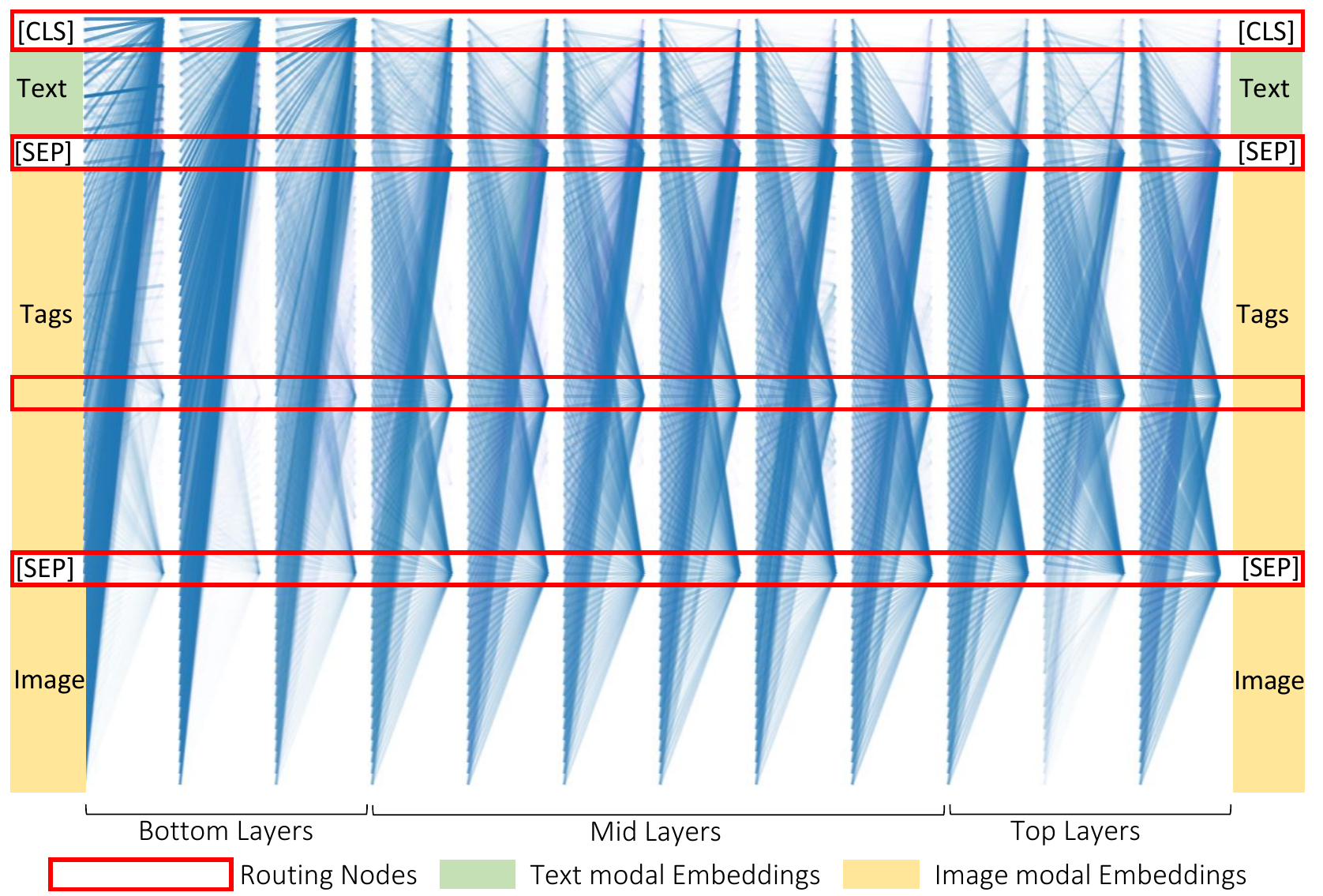} \\
	\end{tabular}
	\caption{Illustration of the ``routing nodes'' inside each layers of the VL transformer VinVL~\cite{vinvl} that receive most of the attention (better viewed with zoom in). $\mathtt{[CLS]}$ and $\mathtt{[SEP]}$ are constantly routing nodes in different samples. }
	\label{fig.route}
	\vspace{-10pt}
\end{figure}
To study the dataflow inside a VL transformer, we first conduct a visualization of the attention computation of pre-trained VinVL-base~\cite{vinvl}\footnote{The phenomenon is also observed in Uniter~\cite{uniter}, see appendix C for more cases.}. The visualized samples are from the COCO 1k test set and depicted by the VIG tool~\cite{vig-2019-multiscale}. As the attention map case illustrates in Fig.~\ref{fig.route}, there are four nodes that receive the most attention weights. Among them, three nodes belongs to the special tokens, i.e., $\mathtt{[CLS]}$ or $\mathtt{[SEP]}$, which do not have any modality properties. Meanwhile, the ``routing nodes'' are fixed in each layer. Therefore, it would be possible to decompose the vision and language inputs while keeping these paths unbroken.

\begin{figure}
	\centering
	\small
	\setlength{\tabcolsep}{0.1em}
	\begin{tabular}{cc}
		\includegraphics[width=0.5\linewidth,height=0.37\linewidth]{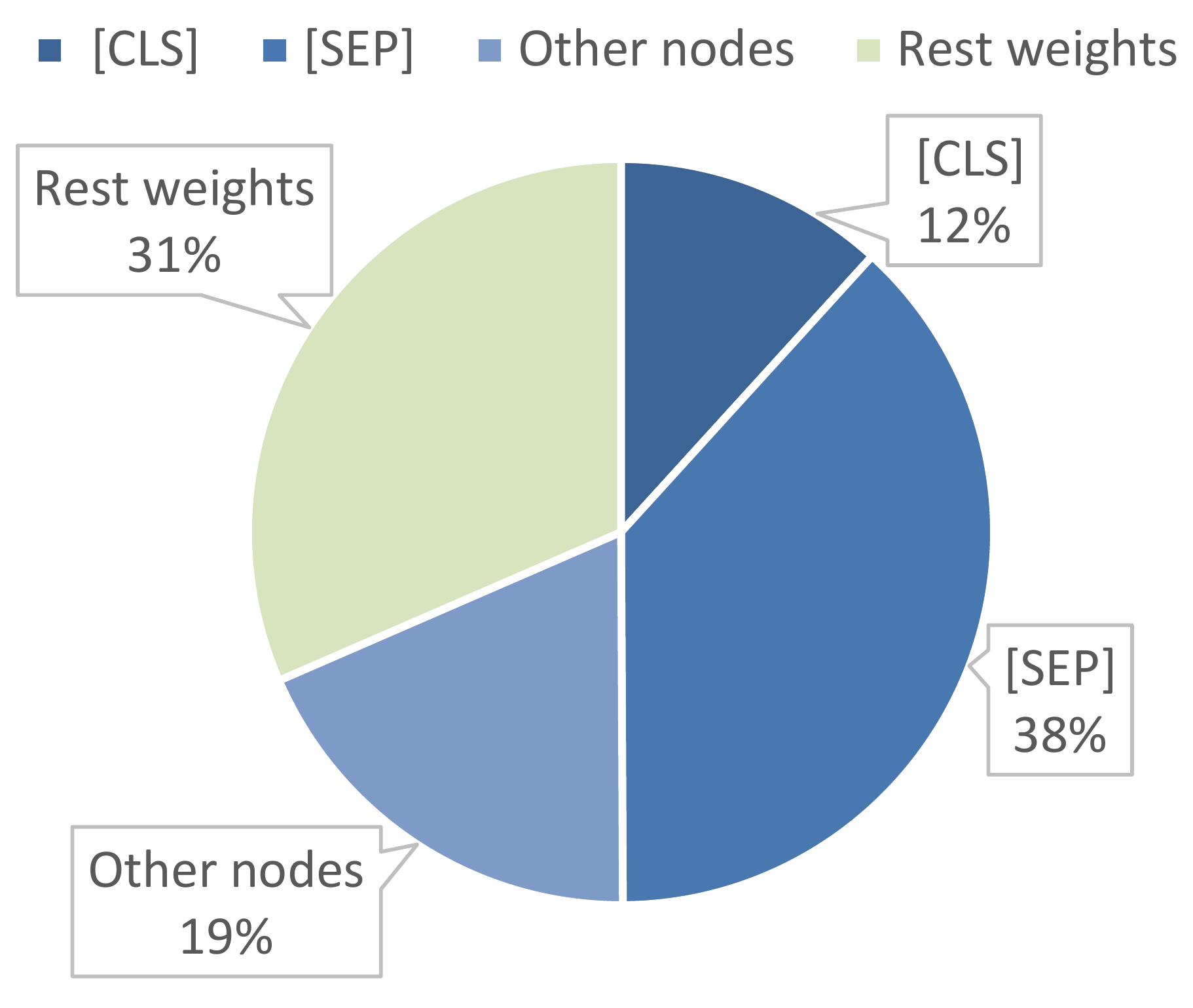} & 
		\includegraphics[width=0.5\linewidth,height=0.37\linewidth]{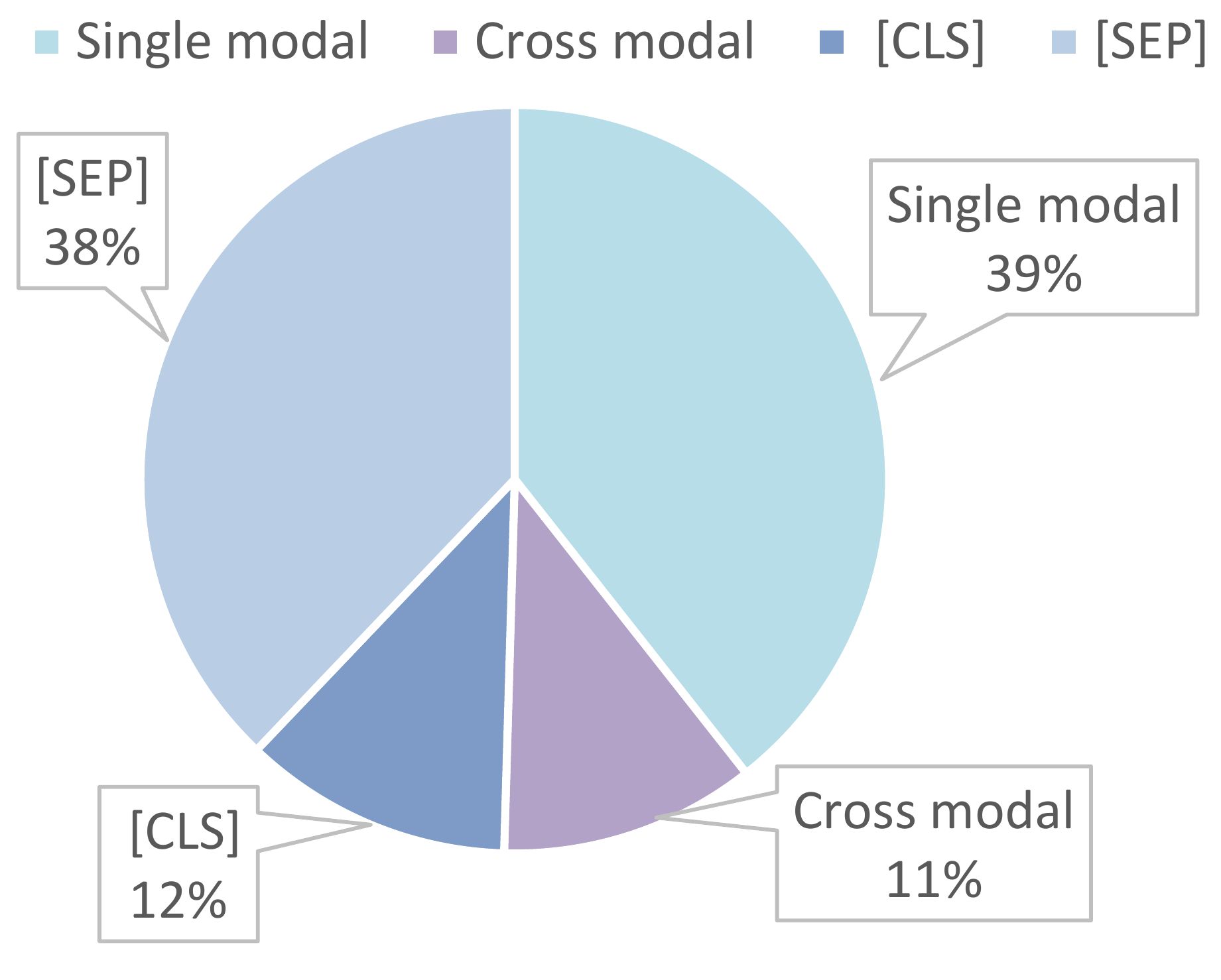} \\
		\footnotesize (a) Percentage of different nodes &
		\footnotesize (b) Percentage of attention types
	\end{tabular}
	\caption{The proportion of attention weights in VinVL according to routing nodes (a) and modality (b) on COCO test set. Only 11\% weights are cross-modal attention.}
	\label{fig.attention_percent_vinvl}
\end{figure}

We collected $1k$ samples and recorded the attention percentage according to modality and the paths to verify this phenomenon. 
To distinguish, we define the following:
\begin{itemize}
	\item Routing node: The top-$k$ tokens\footnote{The value $k$ varies from different VL Transformers, e.g. $k=4$ for VinVL and $k=2$ for Uniter.} that significantly take more proportion of the attention weights than others. 
	\item Neutral attention: The attention that starts from or points to the special tokens $\mathtt{[CLS]}$ and $\mathtt{[SEP]}$ that do not belongs to any modalities. 
	\item Single modal attention: The attention that points from text embeddings to text embeddings or image embedding to image embeddings.
	\item Cross-modal attention: The attention that points to text embeddings from image embeddings or vice versa.
\end{itemize}
As is shown in Fig.~\ref{fig.attention_percent_vinvl} (a), $69$\% of attention weights are on the routing nodes, and half of the attention weights are undertaken by the neutral nodes. Other nodes take a total of $19$\% of the weights.  If the routing nodes are kept after decomposition, the pre-training knowledge is likely to be maintained.
From Fig.~\ref{fig.attention_percent_vinvl} (b) we can see only $11$\% of the attention weights are immediate cross-modal interactions; the remaining $89$\% are not for immediate cross-modal interaction. If the rest of $89$\% attention dataflow is not broken after decomposition, we will only need small data to reconstruct the cross-modal attention and maintain most of the pre-training knowledge.

\begin{table}[]
	\centering
	\small
	\setlength{\tabcolsep}{0.4em}
	\begin{tabular}{@{}lrrrrc@{}}
		\toprule
		\multirow{2}{*}{Layers} & \multicolumn{3}{c}{Neutral (\%)} & \multicolumn{1}{c}{\multirow{2}{*}{Single (\%)}} & \multicolumn{1}{c}{\multirow{2}{*}{Cross (\%)}} \\ \cmidrule(lr){2-4}
		& {[}CLS{]} & {[}SEP{]} & Total & \multicolumn{1}{c}{} & \multicolumn{1}{c}{} \\ \midrule
		Bottom (1-3) & 39.1 & 9.6 & 48.7 & 38.7 & 12.6 \\
		Mid (4-9) & 2.6 & 46.0 & 48.6 & 40.7 & 10.6 \\
		Top (10-12) & 1.5 & 49.1 & 50.6 & 38.6 & 10.8 \\ \bottomrule
	\end{tabular}
	\caption{The attention type proportions of each VinVL layers. Only about 10\% are immediate cross-modal attention.}
	\label{tab.attention_percent_vinvl}
\end{table}
Table~\ref{tab.attention_percent_vinvl} shows the percentage of different attention types in each layer. We can see that single modal attention constantly takes more attention weights than cross-modal attention weights. $\mathtt{[CLS]}$ receives more attention in bottom layers, while on the mid and top layers, $\mathtt{[SEP]}$ becomes more important. Therefore, using $\mathtt{[SEP]}$ or the average of the layer outputs as the representation vector is likely to be more effective than the common practice to use $\mathtt{[CLS]}$.

\begin{figure*}
	\centering
	\includegraphics[width=\linewidth]{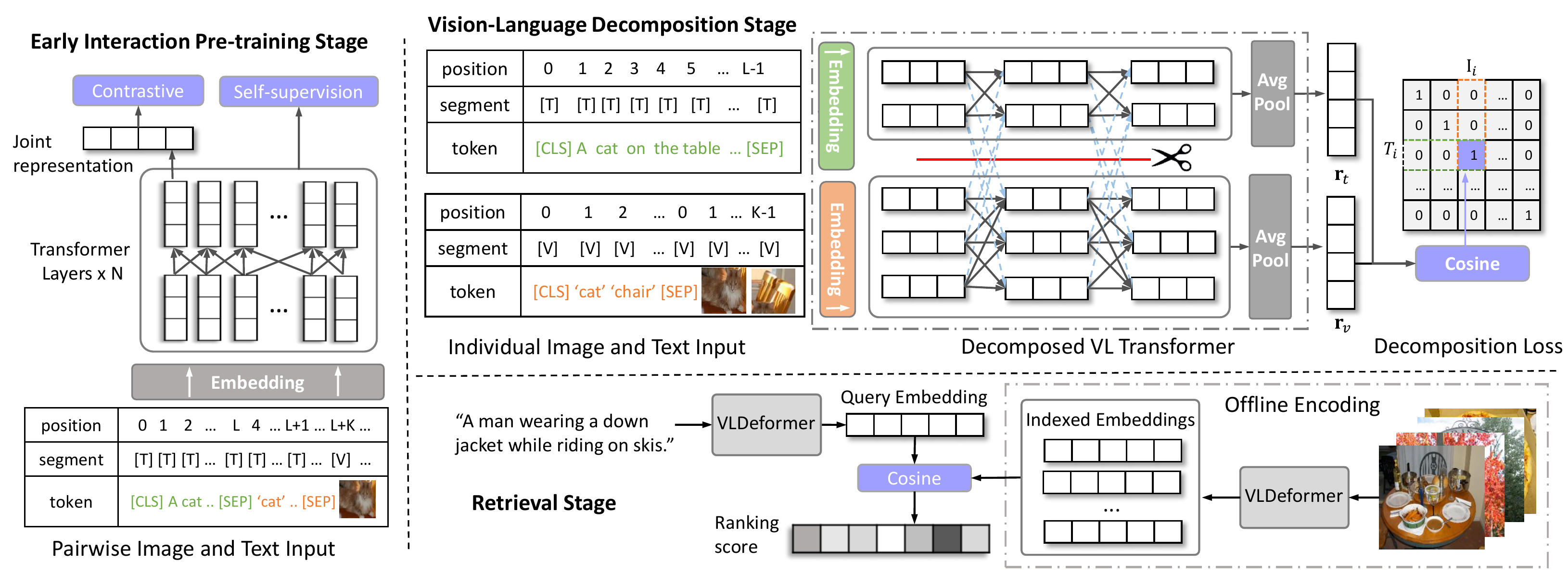}
	\caption{Overall structure of the Vison-Language Transformer Decomposing (VLDeformer). The cross-modal attention inside the transformer module is decomposed during decomposition stage (illustrated by the red line with the scissors).}
	\label{fig:framework}
	\vspace{-5pt}
\end{figure*}

\section{Methodology}
\label{sec.architecture}
The paradigm of the VLDeformer is illustrated in Fig.~\ref{fig:framework}. It includes early-interaction pre-training, vision-language decomposing, and retrieval stages. Since there are many ways~\cite{unicoder,uniter,imagebert,vinvl} to achieve the pre-training stage, in this section, we mainly elaborate on the principles for vision-language decomposing.

\subsection{Early-interaction Pre-training}
The early-interaction pre-training plays the role of learning the cross-modal alignment from large-scale datasets. To exploit the fine-grained relationship between text and image modality, the pairwise text and image input are concatenated and fed to the network simultaneously.

$\textbf{Pairwise Image and Text Input}$
The pairwise text and image input includes position embedding, segment embedding, and token embedding. The input text $T$ is tokenized as a token sequence $\{w_1,...w_L\}$ where $L$ is the length of the WordPiece \cite{wordpiece} tokenizer output. The input image $I$ is pre-processed by the object detection network~\cite{vinvl} to extract region features and tags. As for the segment tokens, we assign $\mathtt{[T]}$ segment token to mark the word tokens and the object tags, and $\mathtt{[V]}$ to represent the region features. The final embedding for both text and image input is obtained by the summing up of position, segment, and token embedding, followed by a layer normalization.  

The text and image features interact through self-attention in the network. 
The pre-training objectives are self-supervision tasks, i.e., mask language modeling, and contrastive learning for cross-modal alignment on the joint representation. Since most of the existing VL transformers are built in this pre-training procedure~\cite{unicoder,uniter,imagebert,vinvl}, VLDeformer can be applied to any of them. In this section the model is trained as a VinVL-base~\cite{vinvl} model.

\subsection{Vision-language Decomposition}
\textbf{Individual Image and Text Input}
\label{sec.input}
The concatenated text and image input are divided for individual encoding. 
Guided by the pilot analysis, there are two differences in the format: the special tokens and the position embedding. The special $\mathtt{[CLS]}$ token is added to the beginning of both modalities, and $\mathtt{[SEP]}$ is added to the end of the text and tag tokens. 
The position index for tokens, tags, and objects are assigned separately to distinguish the modality. The text position index ranges from $0$ to $L-1$, while for the image input, the position index starts again from $0$ to $K-1$ where $K$ is the number of the objects or tags. 

\textbf{Decomposed VL Transformer}
The individual image and text input isolates the cross-modal interaction of the VL transformer. To keep the other interaction as much as possible, the network shares weights for the text and image modality.
According to the analysis of the pilot experiment, the $\mathtt{[CLS]}$ node receives small attention weights in the top layers of VL transformer, which is in conflict with the common practice to use $\mathtt{[CLS]}$ as representation. Therefore we experiment with three kinds of representations: $\mathtt{[CLS]}$, $\mathtt{[SEP]}$ and average pooling of all the outputs. 
In practice, we find the average pooling and $\mathtt{[SEP]}$ representation are more effective than the $\mathtt{[CLS]}$ vector for representation, while average pooling representation is slightly better than $\mathtt{[SEP]}$ (see details in Sec.~\ref{sec.ablation}). Finally, the representations $\mathbf{r}_t$ or $\mathbf{r}_v$ are obtained by the average pooling layer with $\tanh$ activation.

\textbf{Decomposition Loss}
\label{sec.optimization}
The objective of decomposition is to reconstruct the broken cross-modal interactions and learn cross-modal similarity through the late-interaction dataflow. We experiment with BCE loss, Triplet loss, and infoNCE loss, with infoNCE loss ultimately achieving the best performance (see details in Sec.~\ref{sec.ablation}).
The infoNCE loss minimizes the cosine distance between semantically aligned samples and maximizes the distance between dissimilar samples.
In a mini-batch with $N$ text-image pairs, we regard the aligned pairs as the positive samples and other combinations as the negatives.
We use an objective as Eq.~\ref{equ:contrastive_t} to pull semantically close images representation $\mathbf{r}_v^j$ to the text representation $\mathbf{r}_t^i$ and push non-close samples apart:
\begin{equation}
\label{equ:contrastive_t}
\mathcal{L}_c^t=-\log \frac{e^{\cos (\mathbf{r}_t^i,\mathbf{r}_v^i)/\tau}}
{\Sigma^N_{j=1} e^{\cos (\mathbf{r}_t^i,\mathbf{r}_v^{j})/\tau}}
\end{equation}
where $\tau$ is a temperature hyper-parameter, and $\cos$ is the cosine similarity $\frac{\mathbf{r}^\mathsf{T}_t\mathbf{r}_v}{\Vert\mathbf{r}_t\Vert \cdot \Vert\mathbf{r}_v\Vert}$. The $\mathcal{L}_c^t$ term can also be regarded as optimizing text-to-image retrieval in a mini-batch.

Symmetric to $\mathcal{L}_c^t$, we use the loss term as in Eq.~\ref{equ:contrastive_v} to learn the image-to-text condition.
\begin{equation}
\label{equ:contrastive_v}
\mathcal{L}_c^v=-\log \frac{e^{\cos (\mathbf{r}_v^i,\mathbf{r}_t^i)/\tau}}
{\Sigma^N_{j=1} e^{\cos (\mathbf{r}_v^i,\mathbf{r}_t^{j})/\tau }}
\end{equation}
The complete contrastive learning loss is the summing up of these two terms:
\begin{equation}
\label{equ:contrastive_total}
	\mathcal{L}_c = \mathcal{L}_c^t + \mathcal{L}_c^v
\end{equation}

Since the main goal of this paper is to show the pre-training and decomposition paradigm, we find that the simple infoNCE loss is enough to maintain comparable performance to the VL transformer.  Other self-supervision could also be useful, which will be left for future work.

\subsection{VLDeformer based Cross-modal Retrieval}
VLDeformer is an individual encoder and therefore enables encoding the retrieval contents offline. For example, in text-to-image retrieval, the images are encoded to embeddings offline so that the online computation only includes the query encoding and the cosine similarity, which is the main reason to achieve the retrieval speed acceleration. 

In this part, we take text-to-image retrieval as an example to introduce the retrieval process. To formulate, the image set is denoted as $\{I_i\}_{i=1}^N$ where $N$ is the image set size. The query is denoted as $T$.

In the offline encoding stage, the images are processed following Sec.~\ref{sec.input} to get the position, segment, and token embeddings and then passed to the VLDeformer model to get the image embedding $\{\mathbf{r}_v^i\}_{i=1}^N $ . The image embeddings could be reused to compare with each text query.

During online retrieval, the query text is first processed to position, segment, and token embeddings then encoded into query embedding $\mathbf{r}_t$. The index of top-$1$ related images to the query text is calculated as Eq.~\ref{equ.hasshing}:
\begin{equation}
\label{equ.hasshing}
n = \mathop{\arg\max}_{ i \in [0, N]} \rm cos(\mathbf{r}_v^i, \mathbf{r}_t) \\
\end{equation}
The top-$1$ retrieved images $I_n$ could be obtained from the image set.

\subsection{Implementation Details}
All the processed images are first resized to $256 \times 256$, then $50$ region-of-interests and the object tags are extracted. The max sequence length of text tokens is set to $35$. The batch size for contrastive decomposing is set to $1750$, while the temperature is set to $0.005$. The AdamW optimizer is adopted with a learning rate of $5e^{-5}$ and weight decay of $1e^{-4}$. The model is trained on an NVIDIA DGX with a Ubuntu 18.04 system and 8 V100 GPU. 
\section{Experiments}
\subsection{Datasets and Evaluation Protocols}
\begin{table*}[]
	\centering
	\small
	\setlength{\tabcolsep}{3pt}
	\begin{tabular}{@{}llcccccccccccccccl@{}}
		\toprule
		\multirow{3}{*}{Methods} & \multicolumn{7}{c}{COCO Test (5k images)} &  & \multicolumn{7}{c}{Flickr30k Test (1k images)} & \multirow{3}{*}{\begin{tabular}[c]{@{}c@{}}Pre-train data\\ $\times 10^6$ pairs\end{tabular}} & \multirow{3}{*}{\begin{tabular}[c]{@{}c@{}}Params\\ $\times 10^6$\end{tabular}} \\ \cmidrule(lr){2-16}
		& \multicolumn{3}{c}{Text Retrieval} &  & \multicolumn{3}{c}{Image Retrieval} &  & \multicolumn{3}{c}{Text Retrieval} &  & \multicolumn{3}{c}{Image Retrieval} &  &  \\
		& R@1 & R@5 & R@10 &  & R@1 & R@5 & R@10 &  & R@1 & R@5 & R@10 &  & R@1 & R@5 & R@10 &  &  \\ \hline
		& \multicolumn{16}{c}{Visual-Semantic Embeddings} &  \\
		CAAN\cite{CAAN} & 52.5 & 83.3 & 90.9 &  & 41.2 & 70.3 & 82.9 &  & 70.1 & 91.6 & 97.2 &  & 52.8 & 79.0 & 87.9 & - & 11 \\
		IMRAM\cite{IMRAM} & 53.7 & 83.2 & 91.0 &  & 39.7 & 69.1 & 79.8 &  & 74.1 & 93.0 & 96.6 &  & 53.9 & 79.4 & 87.2 & - & 18 \\
		SGRAF\cite{SGRAF} & 57.8 & - & 91.6 &  & 41.9 & - & 81.3 &  & 77.8 & 94.1 & 97.4 &  & 58.5 & 83.0 & 88.8 & - & 19 \\
		DIME\cite{sigir2021} & 59.3 & 85.4 & 91.9 &  & 43.1 & 73.0 & 83.1 &  & 81.0 & 95.9 & 98.4 &  & 63.6 & 88.1 & 93.0 & - & 116 \\ \hline
		& \multicolumn{16}{c}{Pre-trained Visual-Semantic Embeddings} &  \\
		*ALIGN & 77.0 & 93.5 & 96.9 &  & 59.9 & 83.3 & 89.8 &  & 95.3 & 99.8 & 99.9 &  & 84.9 & 97.4 & 98.6 & 1800 & 900 \\
		ALIGN-small\cite{ALIGN} & 52.0 & - & - &  & 39.2 & - & - &  & - & - & - &  & - & - & - & 180 & 235 \\
		LightningDOT\cite{lightningdot} & 70.0 & 91.6 & 95.5 &  & 54.0 & 80.8 & 88.5 &  & 83.9 & 97.2 & 98.6 &  & 69.9 & 91.1 & 95.2 & 9.5 & 220 \\\midrule
		VLDeformer & \textbf{72.6} & \textbf{91.9} & \textbf{96.2} &  & \textbf{54.9} & \textbf{81.3} & \textbf{88.8} &  & \textbf{93.5} & \textbf{98.7} & \textbf{99.2} &  & \textbf{80.2} & \textbf{95.1} & \textbf{97.8} & 8.8 & 111 \\ \bottomrule
		\multicolumn{18}{l}{*ALIGN is achieved using $200+$ times larger pre-train data than other pre-trained models, so we mainly compare with ALIGN-small.} 
	\end{tabular}
	\vspace{-5pt}
	\caption{Comparison results with Visual-Semantic Embedding Methods on COCO and Flickr30k dataset. VLDeformer outperforms other pre-trained models using similar or even smaller data. }
	\label{tab.cme}
\end{table*}

\begin{table*}[]
\centering
\small
\setlength{\tabcolsep}{3.3pt}
\begin{tabular}{lccclcccllccclccclr} 
\toprule
\multirow{3}{*}{Methods} & \multicolumn{8}{c}{Flickr30k Test (1k images)} &  & \multicolumn{7}{c}{COCO Test (1k images)} & \multicolumn{1}{c}{} & \multicolumn{1}{l}{\multirow{3}{*}{Time (s)}} \\ 
\cmidrule{2-18}
& \multicolumn{3}{c}{Text Retrieval} & \multicolumn{1}{c}{} & \multicolumn{3}{c}{Image Retrieval} &  &  & \multicolumn{3}{c}{Text Retrieval} & \multicolumn{1}{c}{} & \multicolumn{3}{c}{Image Retrieval} &  & \multicolumn{1}{l}{} \\ 
\cmidrule{2-4}\cmidrule{6-8}\cmidrule{11-13}\cmidrule{15-17}
& R@1 & R@5 & R@10 &  & R@1 & R@5 & R@10 & AR &  & R@1 & R@5 & R@10 &  & R@1 & R@5 & R@10 & AR & \multicolumn{1}{l}{} \\ 
\midrule
LightningDOT & 83.9 & 97.2 & 98.6 &  & 69.9 & 91.1 & 95.2 & 89.3 &  & - & - & - &  & - & - & - & - & 2.7 \\
+Reranker~\cite{lightningdot} & 87.2 & 98.3 & 99.0 &  & 75.6 & 94.0 & 96.5 & 91.7 &  & - & - & - &  & - & - & - & - & 37.6 \\
UnicoderVL~\cite{unicoder} & 86.2 & 96.3 & 99.0 &  & 71.5 & 90.9 & 94.9 & 89.8 &  & 84.3 & 97.3 & 99.3 &  & 69.7 & 93.5 & 97.2 & 90.2 & - \\
Uniter~\cite{uniter} & 86.9 & 98.1 & 99.2 &  & 75.5 & 94.0 & 96.6 & 91.7 &  & - & - & - &  & - & - & - & - & 405.3 \\
Oscar-base~\cite{oscar} & - & - & - &  & - & - & - &  &  & 88.4 & 99.1 & 99.8 &  & 75.7 & 95.2 & \textbf{98.3} & 92.7 & 1300.5 \\
VinVL-base~\cite{vinvl} & \textbf{93.6} & \textbf{99.1} & \textbf{99.9 } &  & \textbf{82.0} & \textbf{95.7} & 97.7 & \textbf{94.6} &  & \textbf{89.8} & 98.8 & 99.7 &  & \textbf{78.2 } & \textbf{95.6} & 98.0 & \textbf{93.3} & 1301.5 \\ 
\midrule
VLDeformer & 93.5 & 98.7 & 99.2 &  & 80.2 & 95.1 & \textbf{97.8} & 94.0 &  & 89.2 & \textbf{98.9} & \textbf{99.9} &  & 75.9 & 95.4 & 98.0 & 92.9 & \textbf{1.2} \\
\bottomrule
\end{tabular}
\caption{Cross-modal retrieval comparison results to VL transformers on COCO and Flickr30k dataset. VLDeformer achieves $1000+$ acceleration with less than $0.6$\% average recall drop. }
\vspace{-10pt}
\label{tab.retrieval}
\end{table*}
\textbf{Datasets}
The pre-training stage uses $8.8$M text-image pairs form public datasets as VinVL~\cite{vinvl}. The COCO~\cite{coco} and the Flickr30k~\cite{flickr30k} datasets are used for the decomposition stage. The COCO dataset contains $123$K images and is divided into $114$K training, $5$K validation, and $5$K test images. We also use a common split of $1$k tests for comprehensive evaluation.The Flickr30k dataset contains $31$K images which are divided into 29K/1K/1K for training, validation, and test.  Each image has $5$ caption texts. 

\textbf{Evaluation}
The retrieval performance is measured by the recall at top$k$ samples (R@$k$). Three $k$ values, R@1, R@5, and R@10, are reported for text-to-image retrieval and vice versa. We evaluate the retrieval speed for the text-to-image retrieval using 1k, 5k, and 10k text-image pairs.

\subsection{Retrieval Accuracy Analysis}
\label{sec.retrieval}
\subsubsection{Comparison with Visual-Semantic Embeddings}
Table~\ref{tab.cme} shows the comparison results of VLDeformer and visual-semantic embedding methods. Both VLDeformer and other pre-trained models substantially outperform the models without pre-training like CAAN and DIME. 
It is worth noting that the performance of pre-trained visual-semantic embeddings varies depending on the pre-training data scale. For example, the *ALIGN trained on the largest data outperforms the other models. However, the pre-training dataset of *ALIGN is $204$ times larger than our pre-training dataset and $189$ times larger than that of LightningDOT, making it hard to judge these models fairly. If we compare the models in similar pre-training data scales, the smaller ALIGN-small model trained on $180$M text-image pairs has a dramatic performance drop as the data scale decrease. 

It is worth noting that VLDeformer outperforms all state-of-the-art visual-semantic embedding models when compared to similar or even smaller data sizes. Therefore we can conclude that VLDeformer is the most effective visual-semantic embeddings method on a comparable data scale.

\subsubsection{Comparison with VL Transformers}
Table~\ref{tab.retrieval} shows the retrieval score and time cost comparison between the VLDeformer network and state-of-the-art VL transformers on COCO and Flickr30k 1k test sets. Compared with the backbone VinVL-base model, VLDeformer achieves thousands of acceleration with less than 0.5\% drop in average recall and can even outperform it on R@5 and R@10 levels at COCO 1k text retrieval set. VLDeformer also outperforms other VL transformers like Unicoder-VL and Uniter. Compared with the pre-trained two-branch transformer, LightningDOT, VLDeformer achieves better results in both accuracy and inference time and also outperforms LightningDOT with an Oscar Reranker.

Still, there is a performance gap between VLDeformer and the backbone VinVL model, which is more obvious on the R@1 image retrieval score, i.e., $2.3$\% on COCO and $1.8$\% on Flickr. However, the difference on R@$5$ and R@$10$ is very small, which means that many ground truth images are not hit by the top$1$ result but recalled within top$5$ records. 
\begin{figure*}
	\centering
	\small
	\setlength{\tabcolsep}{0.1em}
	\begin{tabular}{cccccccccc}
		\includegraphics[width=2.2cm,height=2.0cm]{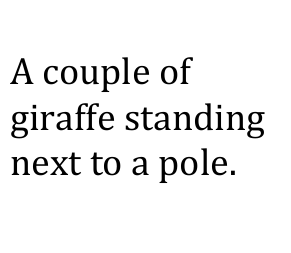} &
		\includegraphics[width=2.2cm,height=2.0cm]{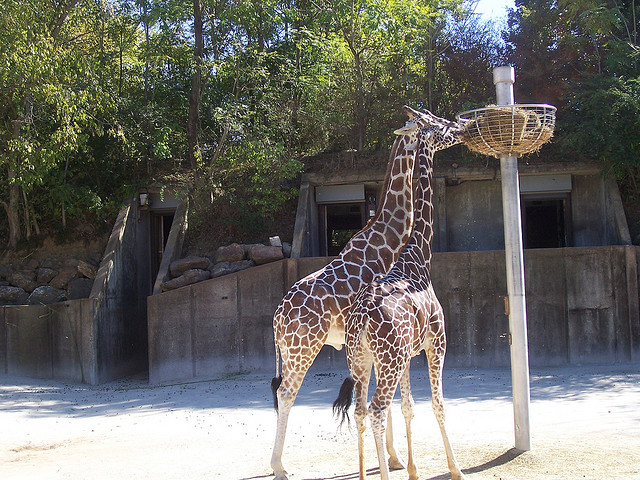} & & &
		\includegraphics[width=2.2cm,height=2.0cm]{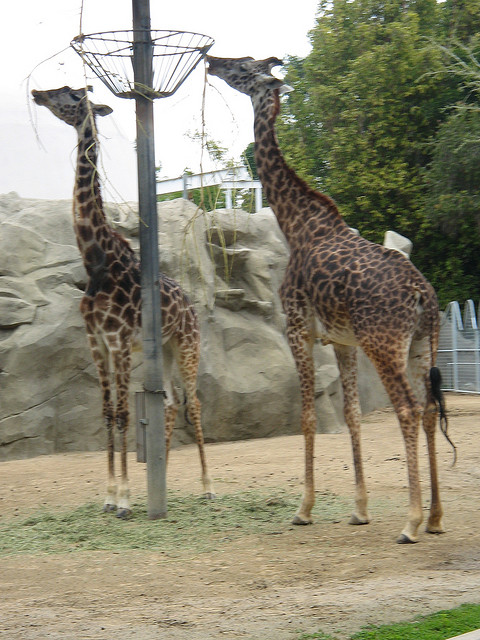} & 
		\includegraphics[width=2.2cm,height=2.0cm]{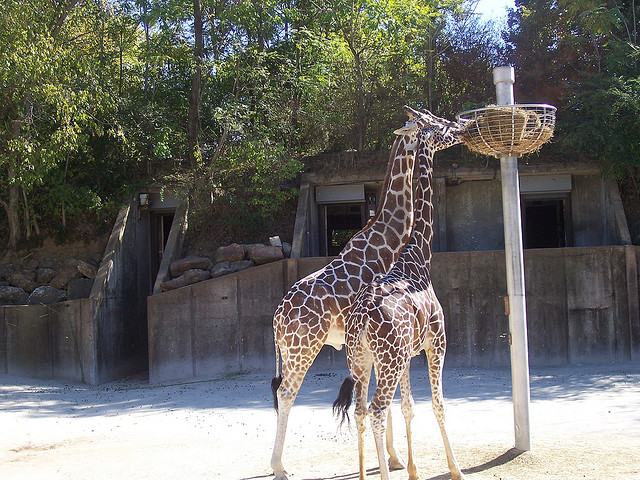} &
		\includegraphics[width=2.2cm,height=2.0cm]{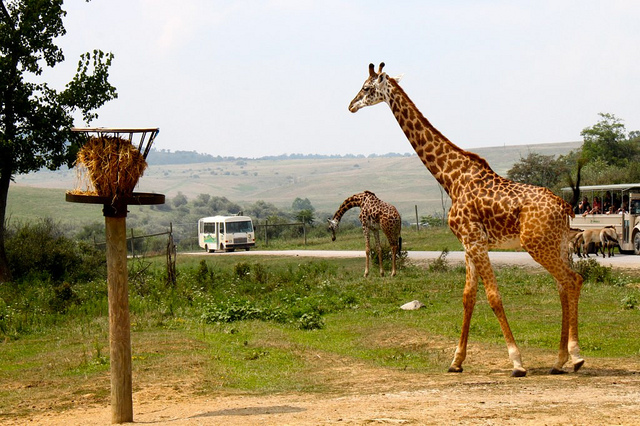} &
		\includegraphics[width=2.2cm,height=2.0cm]{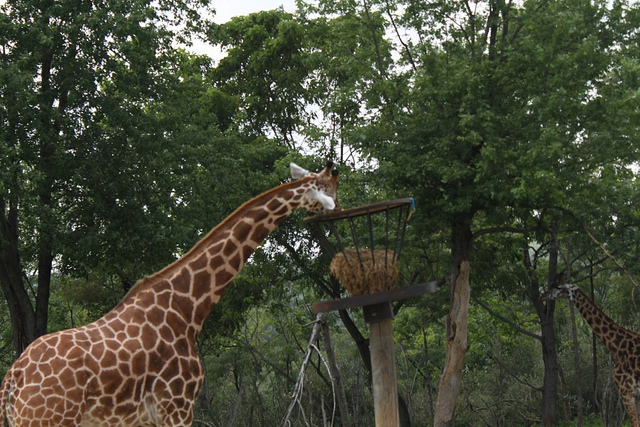} & 
		\includegraphics[width=2.2cm,height=2.0cm]{figures/case_study/t0_i4.png} & \\

		\includegraphics[width=2.2cm,height=2.0cm]{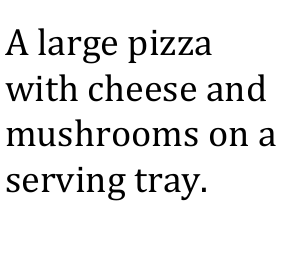} &
		\includegraphics[width=2.2cm,height=2.0cm]{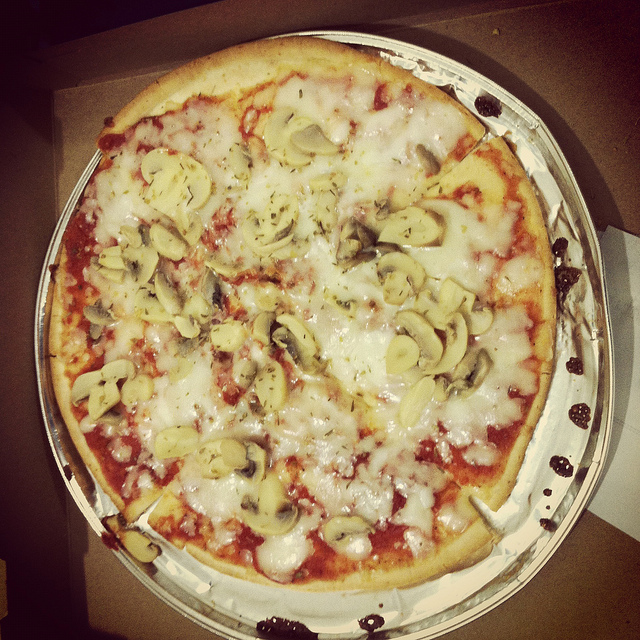} & & &
		\includegraphics[width=2.2cm,height=2.0cm]{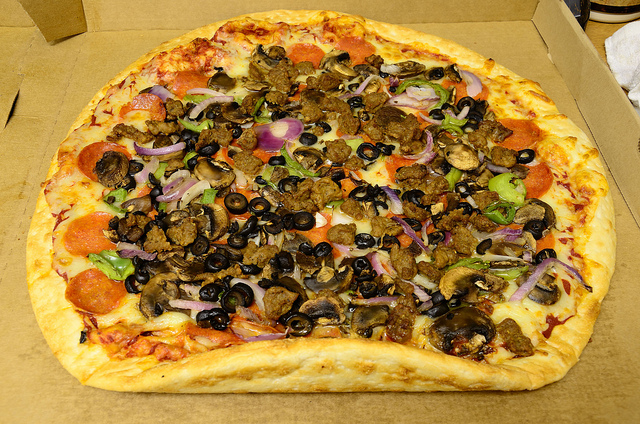} & 
		\includegraphics[width=2.2cm,height=2.0cm]{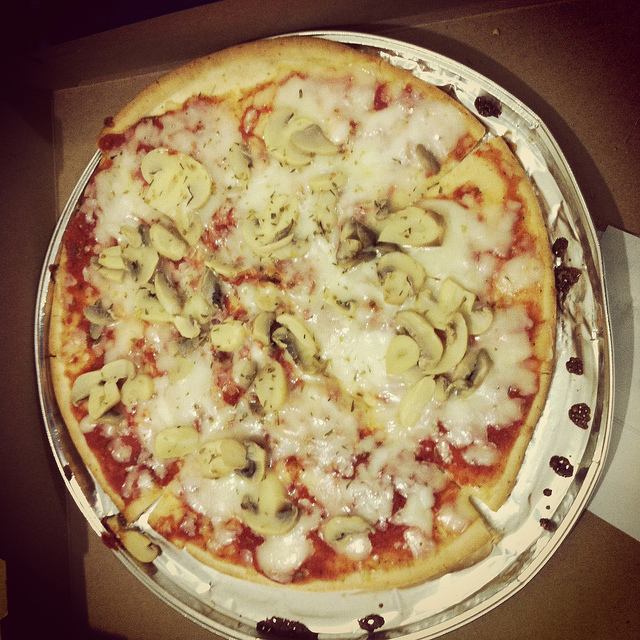} &
		\includegraphics[width=2.2cm,height=2.0cm]{figures/case_study/t3_i2.png} &
		\includegraphics[width=2.2cm,height=2.0cm]{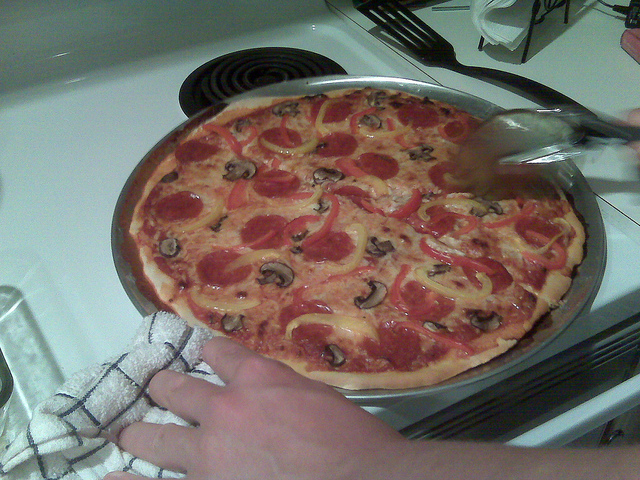} & 
		\includegraphics[width=2.2cm,height=2.0cm]{figures/case_study/t3_i4.png} & \\
		
		\includegraphics[width=2.2cm,height=2.0cm]{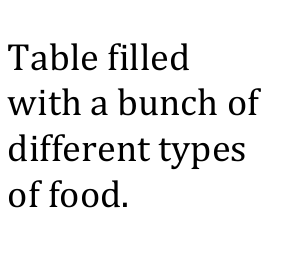} &
		\includegraphics[width=2.2cm,height=2.0cm]{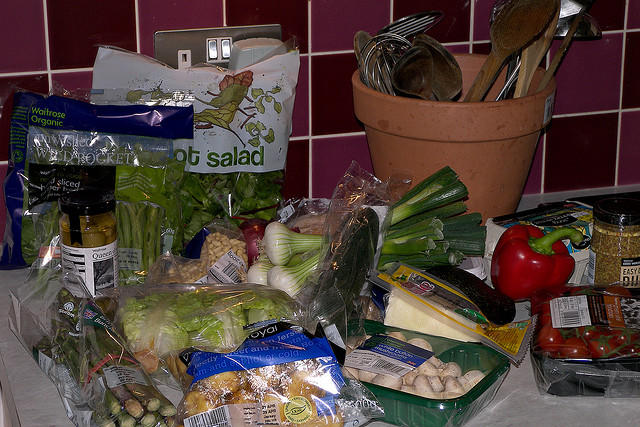} & & &
		\includegraphics[width=2.2cm,height=2.0cm]{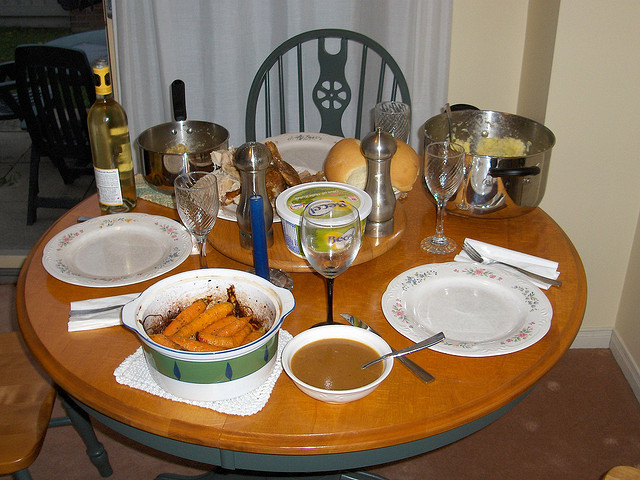} & 
		\includegraphics[width=2.2cm,height=2.0cm]{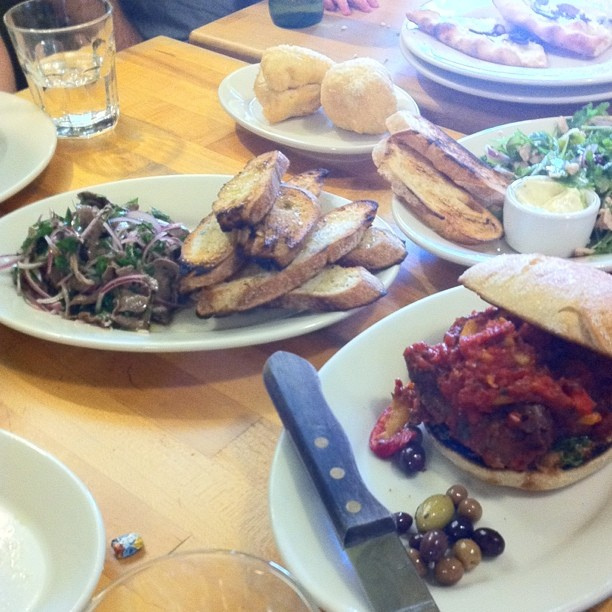} &
		\includegraphics[width=2.2cm,height=2.0cm]{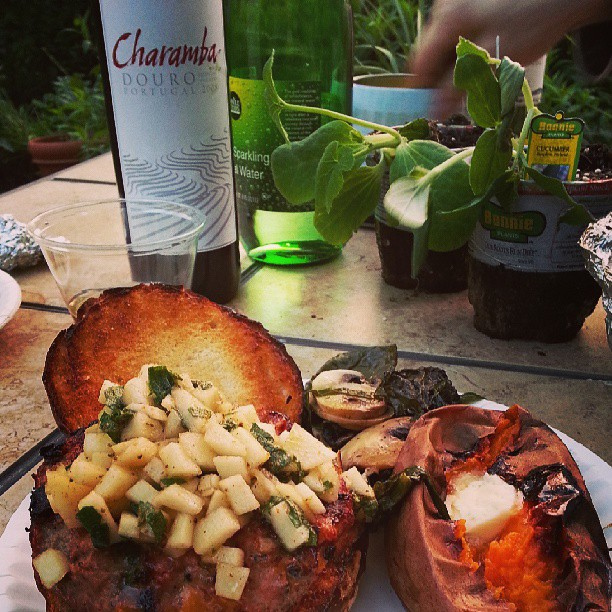} &
		\includegraphics[width=2.2cm,height=2.0cm]{figures/case_study/t6_i3.png} & 
		\includegraphics[width=2.2cm,height=2.0cm]{figures/case_study/t6_i4.png} & \\
		
		\includegraphics[width=2.2cm,height=2.0cm]{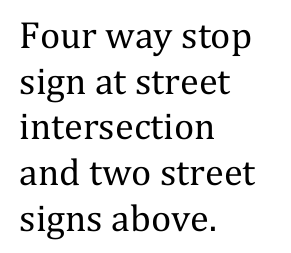} &
		\includegraphics[width=2.2cm,height=2.0cm]{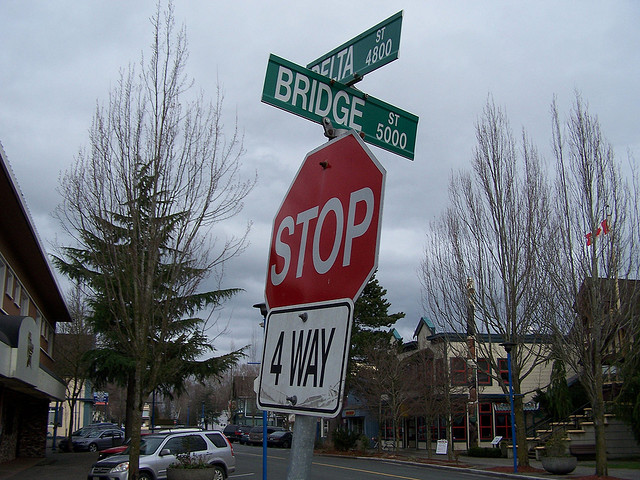} & & &
		\includegraphics[width=2.2cm,height=2.0cm]{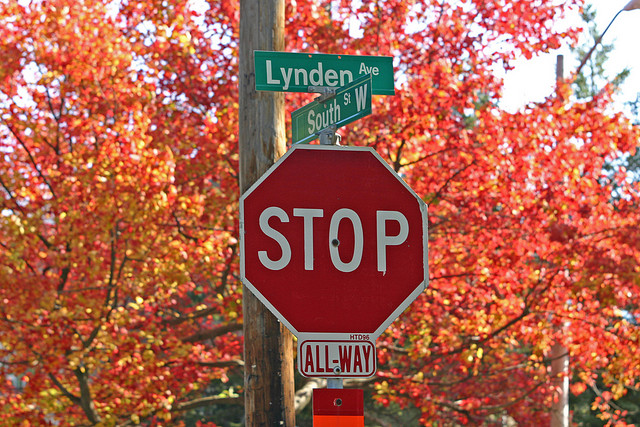} & 
		\includegraphics[width=2.2cm,height=2.0cm]{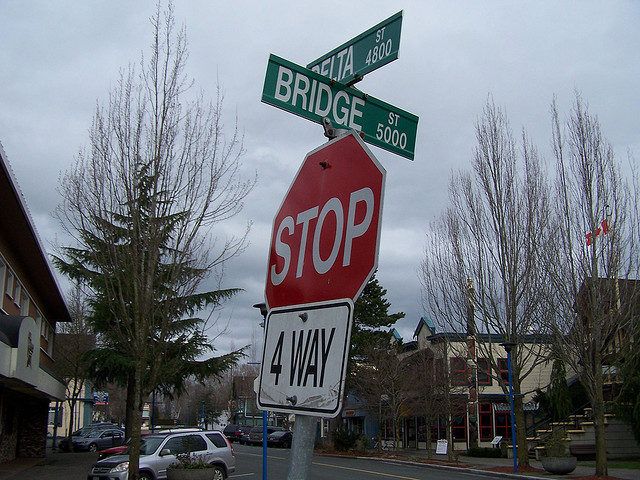} &
		\includegraphics[width=2.2cm,height=2.0cm]{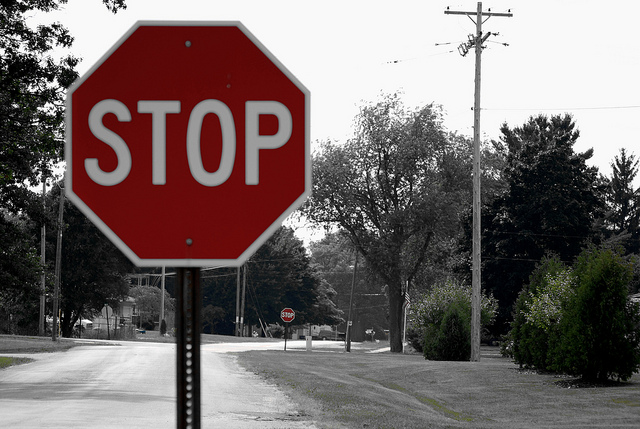} &
		\includegraphics[width=2.2cm,height=2.0cm]{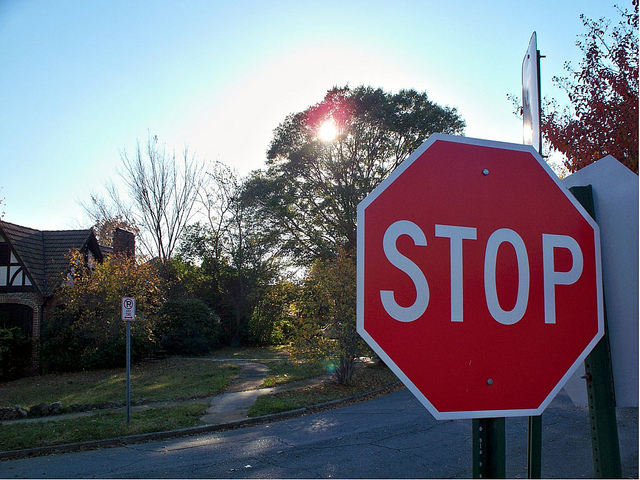} & 
		\includegraphics[width=2.2cm,height=2.0cm]{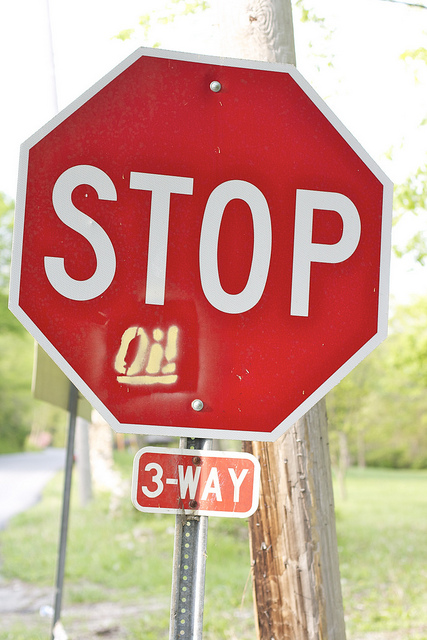} & \\
		
		\includegraphics[width=2.2cm,height=2.0cm]{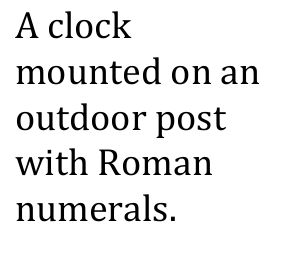} &
		\includegraphics[width=2.2cm,height=2.0cm]{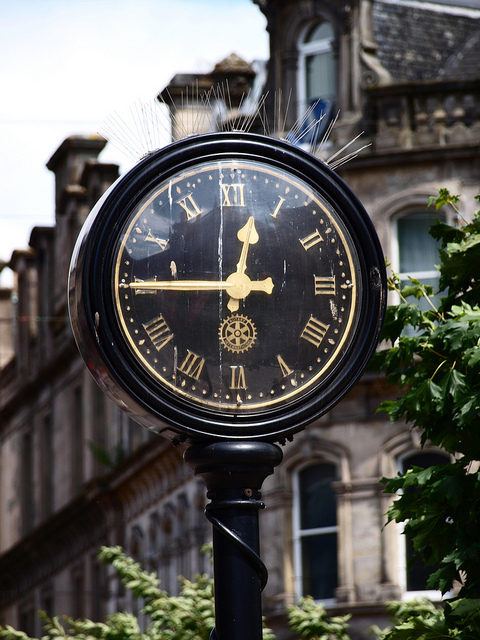} & & &
		\includegraphics[width=2.2cm,height=2.0cm]{figures/case_study/t4_i0.png} & 
		\includegraphics[width=2.2cm,height=2.0cm]{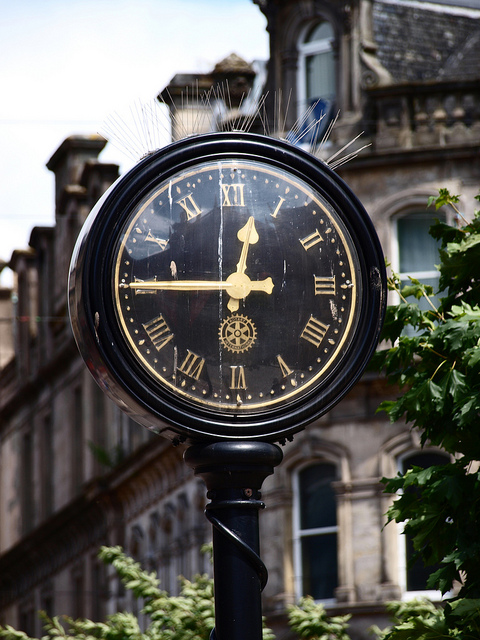} &
		\includegraphics[width=2.2cm,height=2.0cm]{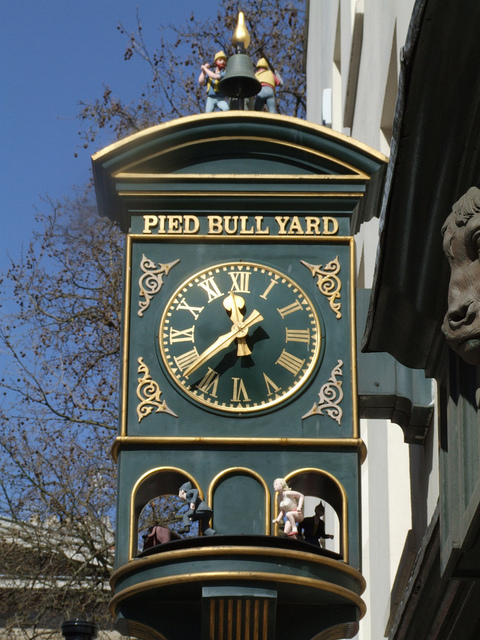} &
		\includegraphics[width=2.2cm,height=2.0cm]{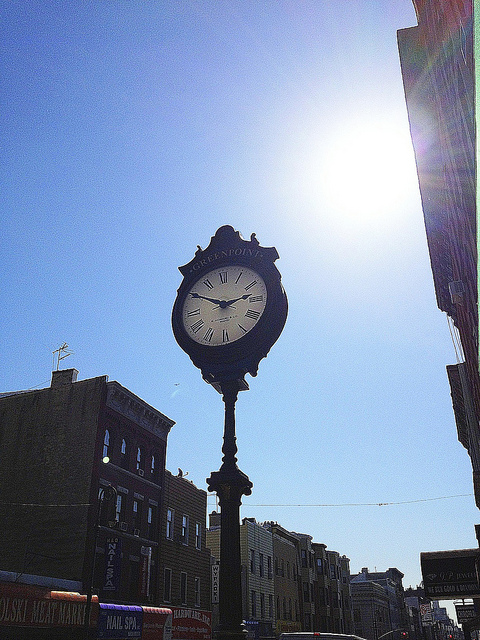} & 
		\includegraphics[width=2.2cm,height=2.0cm]{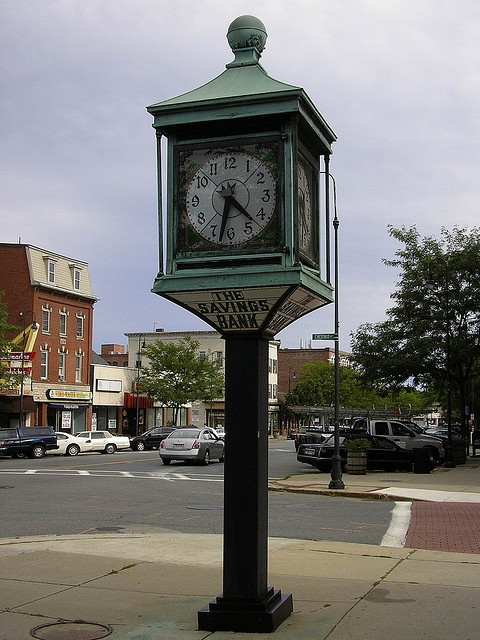} & \\

		Query Text &  Ground Truth  & & & Top@1 & Top@2 & Top@3 & Top@4 & Top@5 &
	\end{tabular}
	\caption{Retrieved top5 images where VLDeformer flips backbone VL Transformer from right to wrong in R@$1$. }
	\vspace{-5pt}
	\label{fig:case}
\end{figure*}

\subsubsection{Qualitative Case Analysis}
\label{sec.case}
Since the R@$1$ metric only calculates the hit ratio of the one aligned ground truth image, it may be inflected by other semantically similar samples. Therefore, we inspect the cases that are properly predicted by the backbone VinVL model at top$1$ but flipped by the VLDeformer. Fig.~\ref{fig:case} shows the top$5$ retrieved images for such cases\footnote{See Appendix B for more detailed qualitative comparisons.}. Interestingly, many images share the same semantics with the query text although they are not the ground truth, e.g, ``two giraffes next to a pole'' or ``four-way stop signs''. For such queries, the top$1$ metric is not applicable to judge the retrieval results. Some queries like the third case have rough semantics that could be aligned with a wide range of images, e.g, “table with different food, e.g, ``table with different food''. These queries also decrease the top$1$ metrics because it is hard to recall the ground truth at top$5$ or even top$10$ records. 
The samples also show some limitations of VLDeformer. For example, the fourth case mainly focuses on the ``clock mounted on outdoor post'' but fails to distinguish the ``roman numeral'' on the dial, indicating that more detailed matching is necessary for future works. 

\subsection{Retrieval Efficiency Analysis}

\begin{figure}
	\centering
	\includegraphics[width=0.95\linewidth]{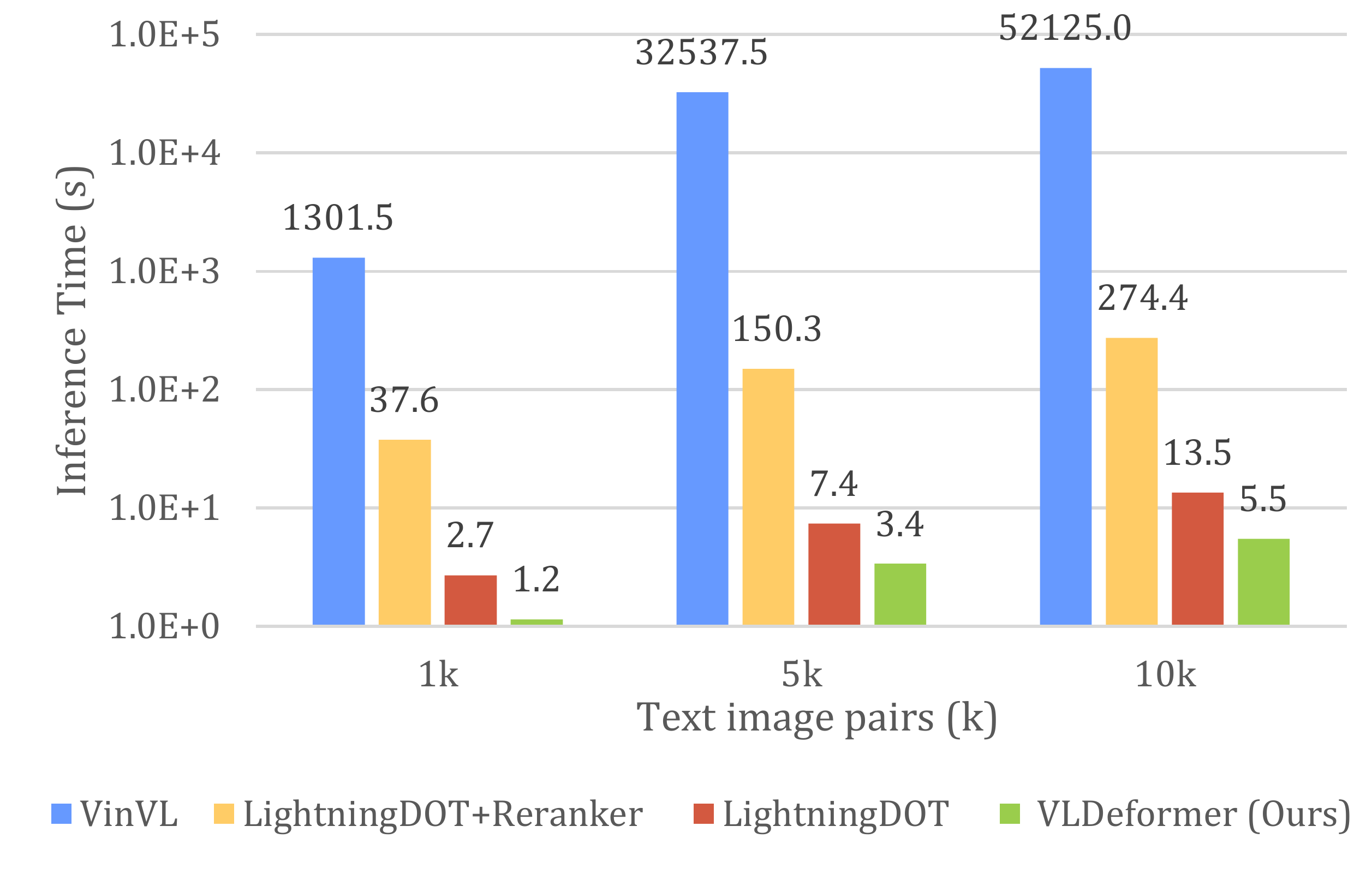}
	\caption{Text-to-image retrieval time on 1k and 5k and 10k image corpus with $400$ mini-batch size. }
	\vspace{-15pt}
	\label{fig:bidirectional_speed}
\end{figure}

The time costs to match all the text-image pairs\footnote{We also report the single query response time in Appendix A.} are shown in Fig.~\ref{fig:bidirectional_speed}. The models are compared on the same machine using one V100 GPU using 400 batch size. Only the inference time is recorded to exclude the data loading time. 
VinVL uses a very long time\footnote{Since VinVL costs a very long time on 5k and 10k, the total time is estimated through the average time cost of the first 1k batches.} (about 0.5 hours on 1k and even more on larger data), and the time response goes like a quadratic function with the data size. In quantitative comparison, VLDeformer achieves more than $1k$ times acceleration on 1k data and $9k$ times on 5k than VinVL. Both VLDeformer and LightningDOT show linear time cost curves as data size increases, but LightningDOT costs more time than VLDeformer, likely because the model is built on a larger BERT-large network. It is worth noting that when LightningDOT uses an Oscar-large~\cite{oscar} reranker to achieve compatible accuracy to VLDeformer, its retrieval time will increase by order of magnitude.

\subsection{Ablation study}
\label{sec.ablation}
To verify the designs for VLDeformer, we conduct an ablation study over the COCO 1k test set. The compared models are trained with the same hyper-parameters. The results are shown in Table~\ref{tab.ablation}.\\
\textbf{Representation selection.}
The results of using  $\mathtt{[CLS]}$ and  $\mathtt{[SEP]}$ as representation are shown in VLDeformer-$\mathtt{[CLS]}$ and VLDeformer-$\mathtt{[SEP]}$. Compared with average pooling representation, VLDeformer-$\mathtt{[SEP]}$ is higher on R@5 and R@10 but lower on R@1, while VLDeformer-$\mathtt{[CLS]}$ decreases significantly on all metrics. The results qualify the observation in Table~\ref{tab.attention_percent_vinvl} that $\mathtt{[CLS]}$ is not as important as $\mathtt{[SEP]}$ at the top layers.
\\
\textbf{Decomposition loss selection.}
To evaluate the effectiveness of the infoNCE loss for decomposition, we compare BCE for pairwise cosine similarity VLDeformer-BCE and triplet loss VLDeformer-triplet. As a result, both of the two objectives perform worse than infoNCE loss, and causes dramatic performance drops, especially on R@$1$.\\
\textbf{Has VLDeformer kept knowledge from the pre-training stage?}
The (w/o pre-training) trains VLDeformer from the vision-language decomposition stage using randomly initialized weights. 
The comparison between VLDeformer and (w/o pre-training) is a black box test. Since the performance drops significantly without the pre-training stage, it can be inferred that some pre-training knowledge is kept after decomposition.
\\
\textbf{Did the decomposition stage reconstruct the pre-trained model?}
Since the cross-modal attention of the pre-trained model is broken in VLDeformer, we wonder to what extent the decomposition reconstructs the pre-trained model. (w/o decompose$^\dag$) shows the performance of directly using the VinVL model as an individual embedding encoder. To our surprise, the scores are very low, indicating that the decomposing stage is necessary to maintain the performance.

\begin{table}
	\centering
	\small
	\setlength{\tabcolsep}{2.7pt}
	\begin{tabular}{@{}lllccccc@{}}
		\toprule
		\multirow{2}{*}{Methods} & \multicolumn{3}{c}{Text Retrieval} &  & \multicolumn{3}{c}{Image Retrieval} \\ \cmidrule(lr){2-4} \cmidrule(l){6-8} 
		& R@1 & R@5 & R@10 &  & R@1 & R@5 & R@10 \\ \midrule
		VLDeformer & \textbf{89.2} & 98.9 & \textbf{99.9} &  & \textbf{75.9} & \textbf{95.4} & 98.0 \\\hline 
		VLDeformer-$\mathtt{[SEP]}$ & 88.5 & \textbf{99.1} & 99.8 &  & 74.9 & 94.8 & \textbf{98.1} \\
		VLDeformer-$\mathtt{[CLS]}$ & 83.0 & 95.3 & 96.6 &  & 69.7 & 87.8 & 90.4 \\ \hline
		VLDeformer-BCE & 72.5 & 95.8 & 99.0 &  & 60.7 & 90.2 & 96.2 \\
		VLDeformer-Triplet & 48.9 & 75.5 & 85.0 &  & 30.6 & 62.3 & 76.4 \\\hline
		w/o pre-training & 81.5 & 97.3 & 98.8 &  & 64.8 & 91.6 & 95.9 \\
		w/o decomposition$^\dag$ & 0.3 & 1.0 & 2.0 &  & 0.1 & 0.2 & 1.6 \\ \bottomrule
	\end{tabular}
	\caption{Ablation study of VLDeformer on COCO 1k Test.	(w/o decomposition$^\dag$ is tested directly using the pre-trained VinVL transformer as an individual encoder.) }
	\vspace{-10pt}
	\label{tab.ablation}
\end{table}

\subsection{Decomposition Analysis}
\begin{table}
\centering
\small
\setlength{\tabcolsep}{0.4em}
\begin{tabular}{cclrrrc}
	\toprule
	\multirow{2}{*}{Model} & \multirow{2}{*}{Modal} & \multirow{2}{*}{Layers} & \multicolumn{3}{c}{Neutral Route (\%)} & Others \\ \cmidrule(lr){4-6}
	& &  & {[}CLS{]} & {[}SEP{]} & total & (\%)  \\ \hline
	
	\multirow{8}{*}{w/ pre-train} & \multirow{4}{*}{V} & Bottom (1-3) & 66.3 & 8.1 & 74.3 & 25.6 \\
	&& Mid (4-9) & 6.1 & 77.9 & 84.0 & 16.0 \\
	&& Top (10-12) & 14.2 & 51.3 & 65.5 & 34.5 \\
	&& Total & 20.5 & 58.1 & 78.6 & 21.3 \\ \cmidrule(lr){2-7}
	&\multirow{4}{*}{L} & Bottom (1-3) & 48.0 & 9.6 & 57.7 & 42.3 \\
	&& Mid (4-9) & 3.5 & 70.6 & 74.1 & 25.9 \\
	&& Top (10-12) & 1.8 & 54.4 & 56.2 & 43.8 \\
	&& Total & 14.2 & 51.3 & 65.6 & 34.4 \\ \hline
	
	\multirow{2}{*}{w/o pre-train} & V & Total & 4.1 & 6.6 & 10.7 & 89.2 \\ \cmidrule(lr){2-7}
	& L & Total & 6.5 & 6.5 & 13.0 & 87.0 \\ \bottomrule
\end{tabular}
\caption{The proportion of different attention weights in each layer of VLDeformer with (w/) and without (w/o) pre-training stage.}
\label{tab.attention_percent_vldeformer}
\end{table}

\begin{figure}
	\centering
	\begin{tabular}{c}
		\includegraphics[width=\linewidth]{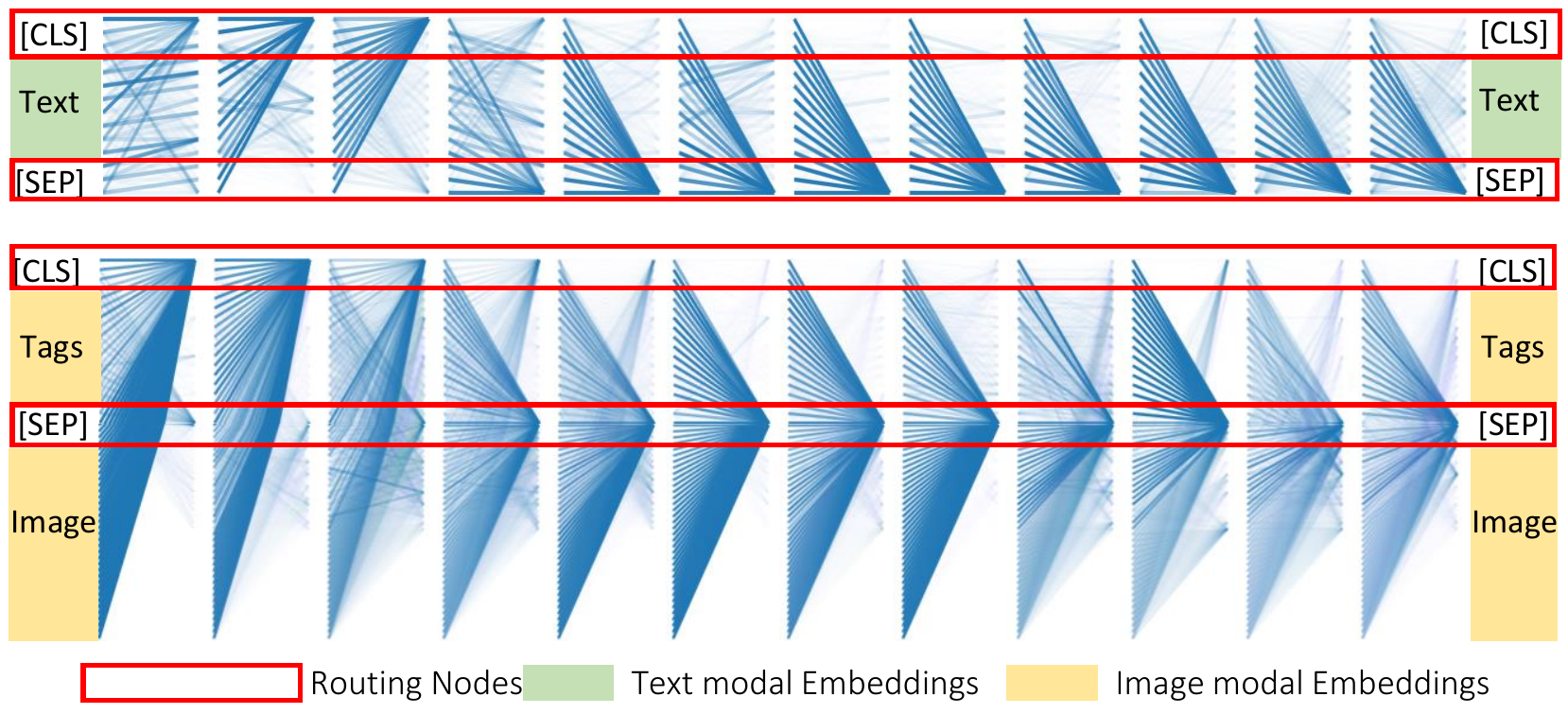} \\
		\footnotesize (a) VLDeformer attention visualization \\
		\includegraphics[width=\linewidth]{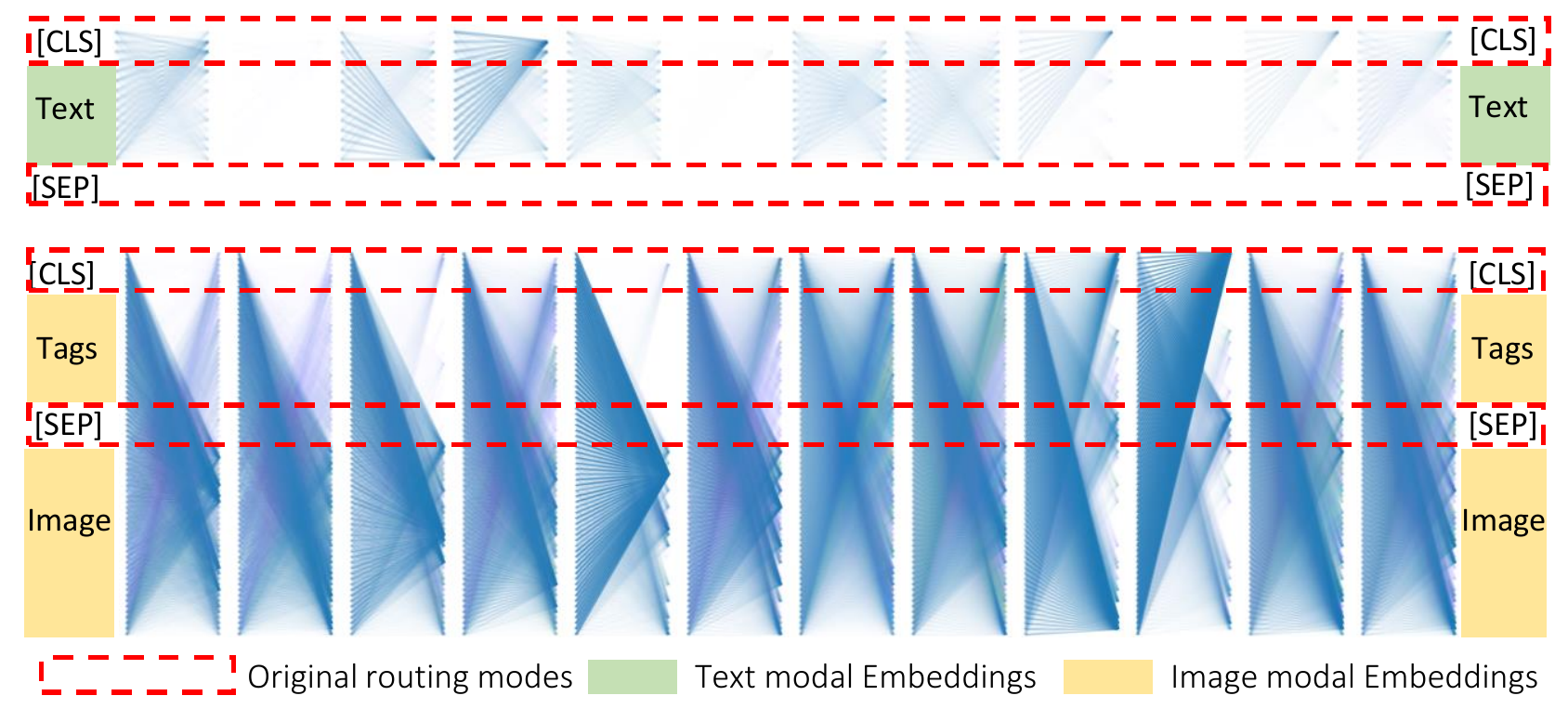} \\
		\footnotesize (b) VLDeformer w/o pre-training attention visualization
	\end{tabular}
	\caption{Attention visualization of VLDeformer with and without pre-training. (From the same sample in Fig.~\ref{fig.route}).}
	\label{fig.attention_nopretrain}
	\vspace{-10pt}
\end{figure}

We further analyze the attention of VLDeformer after decomposition to verify the observation and hypothesis in Sec.~\ref{sec.pilot}. In Table~\ref{tab.attention_percent_vldeformer}, we can find that the $\mathtt{[CLS]}$ and $\mathtt{[SEP]}$ are still important routing nodes which have large proportions of attention weights. In contrast, in the VLDeformer without pre-training, these nodes are not routing nodes. In Fig.~\ref{fig.attention_nopretrain} we compare VLDeformer without pre-train from the same sample\footnote{See more visualization samples in the supplementary files.} as in Fig.~\ref{fig.route}. It can be clearly seen that VLDeformer keeps the routing nodes $\mathtt{[CLS]}$ and $\mathtt{[SEP]}$, but  there is no clear routing node in the VLDeformer without pre-training. 

\section{Conclusion}
VL transformers are effective in cross-modal retrieval but slow in speed. We observed that most of the interactions in VL transformers are not immediate cross-modal attention, but highly rely on neutral nodes.
Therefore we proposed a novel Vision-language Decomposed Transformer (VLDeformer) that pre-trains a VL transformer with early-interaction dataflow and then decomposes it into an individual encoder.  The VLDeformer achieves both $1000+$ times acceleration and less than $0.6$\% average recall drop and also outperforms state-of-the-art visual-semantic embedding models on COCO and Flickr30k datasets.

{\small
	\bibliographystyle{ieee_fullname}
	\bibliography{aaai21}

\begin{thebibliography}{10}\itemsep=-1pt

\bibitem{IMRAM}
Hui Chen, Guiguang Ding, Xudong Liu, Zijia Lin, Ji Liu, and Jungong Han.
\newblock Imram: Iterative matching with recurrent attention memory for
  cross-modal image-text retrieval.
\newblock In {\em 2020 IEEE/CVF Conference on Computer Vision and Pattern
  Recognition (CVPR)}, pages 12652--12660, 2020.

\bibitem{uniter}
Yen-Chun Chen, Linjie Li, Licheng Yu, Ahmed El~Kholy, Faisal Ahmed, Zhe Gan, Yu
  Cheng, and Jingjing Liu.
\newblock Uniter: Universal image-text representation learning.
\newblock In {\em European conference on computer vision}, pages 104--120,
  2020.

\bibitem{bert}
Jacob Devlin, Ming-Wei Chang, Kenton Lee, and Kristina Toutanova.
\newblock Bert: Pre-training of deep bidirectional transformers for language
  understanding.
\newblock In {\em Proceedings of the 2019 Conference of the North American
  Chapter of the Association for Computational Linguistics: Human Language
  Technologies, Volume 1 (Long and Short Papers)}, pages 4171--4186, 2019.

\bibitem{SGRAF}
Haiwen Diao, Ying Zhang, Lin Ma, and Huchuan Lu.
\newblock Similarity reasoning and filtration for image-text matching.
\newblock In {\em Proceedings of the AAAI Conference on Artificial
  Intelligence}, volume~35, pages 1218--1226, 2021.

\bibitem{pfan_6}
Aviv Eisenschtat and Lior Wolf.
\newblock Linking image and text with 2-way nets.
\newblock In {\em Proceedings of the IEEE conference on computer vision and
  pattern recognition}, pages 4601--4611, 2017.

\bibitem{pfan_21}
Kiros Faghri, Fleet and Fidler.
\newblock Vse++: Improving visual-semantic embeddings with hard negatives.
\newblock In {\em BMVC}, page~12, 2018.

\bibitem{pfan_4}
Jiuxiang Gu, Jianfei Cai, Shafiq~R Joty, Li Niu, and Gang Wang.
\newblock Look, imagine and match: Improving textual-visual cross-modal
  retrieval with generative models.
\newblock In {\em Proceedings of the IEEE Conference on Computer Vision and
  Pattern Recognition}, pages 7181--7189, 2018.

\bibitem{ALIGN}
Chao Jia, Yinfei Yang, Ye Xia, Yi-Ting Chen, Zarana Parekh, Hieu Pham, Quoc~V
  Le, Yunhsuan Sung, Zhen Li, and Tom Duerig.
\newblock Scaling up visual and vision-language representation learning with
  noisy text supervision.
\newblock {\em arXiv preprint arXiv:2102.05918}, 2021.

\bibitem{wordpiece}
Melvin Johnson, Mike Schuster, Quoc~V Le, Maxim Krikun, Yonghui Wu, Zhifeng
  Chen, Nikhil Thorat, Fernanda Vi{\'e}gas, Martin Wattenberg, Greg Corrado,
  et~al.
\newblock Google’s multilingual neural machine translation system: Enabling
  zero-shot translation.
\newblock {\em Transactions of the Association for Computational Linguistics},
  5:339--351, 2017.

\bibitem{pfan_2}
Kuang-Huei Lee, Xi Chen, Gang Hua, Houdong Hu, and Xiaodong He.
\newblock Stacked cross attention for image-text matching.
\newblock In {\em Proceedings of the European Conference on Computer Vision},
  pages 201--216, 2018.

\bibitem{SCAN}
Kuang-Huei Lee, Xi Chen, Gang Hua, Houdong Hu, and Xiaodong He.
\newblock Stacked cross attention for image-text matching.
\newblock In {\em Proceedings of the European Conference on Computer Vision
  (ECCV)}, pages 201--216, 2018.

\bibitem{unicoder}
Gen Li, Nan Duan, Yuejian Fang, Ming Gong, and Daxin Jiang.
\newblock Unicoder-vl: A universal encoder for vision and language by
  cross-modal pre-training.
\newblock In {\em Proceedings of the AAAI Conference on Artificial
  Intelligence}, volume~34, pages 11336--11344, 2020.

\bibitem{visualbert}
Liunian~Harold Li, Mark Yatskar, Da Yin, Cho-Jui Hsieh, and Kai-Wei Chang.
\newblock Visualbert: A simple and performant baseline for vision and language.
\newblock {\em arXiv preprint arXiv:1908.03557}, 2019.

\bibitem{oscar}
Xiujun Li, Xi Yin, Chunyuan Li, Pengchuan Zhang, Xiaowei Hu, Lei Zhang, Lijuan
  Wang, Houdong Hu, Li Dong, Furu Wei, et~al.
\newblock Oscar: Object-semantics aligned pre-training for vision-language
  tasks.
\newblock In {\em European Conference on Computer Vision}, pages 121--137,
  2020.

\bibitem{coco}
Tsung-Yi Lin, Michael Maire, Serge Belongie, James Hays, Pietro Perona, Deva
  Ramanan, Piotr Doll{\'a}r, and C~Lawrence Zitnick.
\newblock Microsoft coco: Common objects in context.
\newblock In {\em European conference on computer vision}, pages 740--755.
  Springer, 2014.

\bibitem{vilbert}
Jiasen Lu, Dhruv Batra, Devi Parikh, and Stefan Lee.
\newblock Vilbert: Pretraining task-agnostic visiolinguistic representations
  for vision-and-language tasks.
\newblock {\em arXiv preprint arXiv:1908.02265}, 2019.

\bibitem{pfan_17}
Hyeonseob Nam, Jung-Woo Ha, and Jeonghee Kim.
\newblock Dual attention networks for multimodal reasoning and matching.
\newblock In {\em Proceedings of the IEEE conference on computer vision and
  pattern recognition}, pages 299--307, 2017.

\bibitem{flickr30k}
Bryan~A Plummer, Liwei Wang, Chris~M Cervantes, Juan~C Caicedo, Julia
  Hockenmaier, and Svetlana Lazebnik.
\newblock Flickr30k entities: Collecting region-to-phrase correspondences for
  richer image-to-sentence models.
\newblock In {\em Proceedings of the IEEE international conference on computer
  vision}, pages 2641--2649, 2015.

\bibitem{imagebert}
Di Qi, Lin Su, Jia Song, Edward Cui, Taroon Bharti, and Arun Sacheti.
\newblock {Imagebert: Cross-modal pre-training with large-scale weak-supervised
  image-text data}.
\newblock {\em arXiv}, pages 1--12, 2020.

\bibitem{sigir2021}
Leigang Qu, Meng Liu, Jianlong Wu, Zan Gao, and Liqiang Nie.
\newblock Dynamic modality interaction modeling for image-text retrieval.
\newblock In {\em Proceedings of the 44th International ACM SIGIR Conference on
  Research and Development in Information Retrieval}, pages 1104--1113, 2021.

\bibitem{CLIP}
Alec Radford, Jong~Wook Kim, Chris Hallacy, Aditya Ramesh, Gabriel Goh,
  Sandhini Agarwal, Girish Sastry, Amanda Askell, Pamela Mishkin, Jack Clark,
  et~al.
\newblock Learning transferable visual models from natural language
  supervision.
\newblock {\em arXiv preprint arXiv:2103.00020}, 2021.

\bibitem{vlbert}
Weijie Su, Xizhou Zhu, Yue Cao, Bin Li, Lewei Lu, Furu Wei, and Jifeng Dai.
\newblock Vl-bert: Pre-training of generic visual-linguistic representations.
\newblock In {\em International Conference on Learning Representations}, 2019.

\bibitem{lightningdot}
Siqi Sun, Yen-Chun Chen, Linjie Li, Shuohang Wang, Yuwei Fang, and Jingjing
  Liu.
\newblock Lightningdot: Pre-training visual-semantic embeddings for real-time
  image-text retrieval.
\newblock In {\em Proceedings of the 2021 Conference of the North American
  Chapter of the Association for Computational Linguistics: Human Language
  Technologies}, pages 982--997, 2021.

\bibitem{efficientnet}
Mingxing Tan and Quoc Le.
\newblock Efficientnet: Rethinking model scaling for convolutional neural
  networks.
\newblock In {\em International Conference on Machine Learning}, pages
  6105--6114. PMLR, 2019.

\bibitem{vig-2019-multiscale}
Jesse Vig.
\newblock A multiscale visualization of attention in the transformer model.
\newblock In {\em Proceedings of the 57th Annual Meeting of the Association for
  Computational Linguistics: System Demonstrations}, pages 37--42, Florence,
  Italy, 2019. Association for Computational Linguistics.

\bibitem{pfan_13}
Liwei Wang, Yin Li, Jing Huang, and Svetlana Lazebnik.
\newblock Learning two-branch neural networks for image-text matching tasks.
\newblock {\em IEEE Transactions on Pattern Analysis and Machine Intelligence},
  41(2):394--407, 2018.

\bibitem{pfan++}
Yaxiong Wang, Hao Yang, Xiuxiu Bai, Xueming Qian, Lin Ma, Jing Lu, Biao Li, and
  Xin Fan.
\newblock Pfan++: Bi-directional image-text retrieval with position focused
  attention network.
\newblock {\em IEEE Transactions on Multimedia}, pages 1--1, 2020.

\bibitem{vinvl}
Pengchuan Zhang, Xiujun Li, Xiaowei Hu, Jianwei Yang, Lei Zhang, Lijuan Wang,
  Yejin Choi, and Jianfeng Gao.
\newblock Vinvl: Revisiting visual representations in vision-language models.
\newblock In {\em Proceedings of the IEEE/CVF Conference on Computer Vision and
  Pattern Recognition}, pages 5579--5588, 2021.

\bibitem{CAAN}
Qi Zhang, Zhen Lei, Zhaoxiang Zhang, and Stan~Z Li.
\newblock Context-aware attention network for image-text retrieval.
\newblock In {\em Proceedings of the IEEE/CVF Conference on Computer Vision and
  Pattern Recognition}, pages 3536--3545, 2020.

\bibitem{pfan_19}
Zhedong Zheng, Liang Zheng, Michael Garrett, Yi Yang, Mingliang Xu, and Yi-Dong
  Shen.
\newblock Dual-path convolutional image-text embeddings with instance loss.
\newblock {\em ACM Transactions on Multimedia Computing, Communications, and
  Applications (TOMM)}, 16(2):1--23, 2020.

\end{thebibliography}
}
\end{document}


\title{Supplementary Materials}  

\maketitle
\thispagestyle{empty}
\appendix

\section{Single Query Response Time}
We also record the time of a single query response which calculates the similarity between one text query and all the images. We calculate the P95 and p99.99 percentile values for the single response time on 1k query sentences and 10k images.

The image embeddings of VLDeformer are computed offline as the real-world retrieval scenario. As is shown in Table~\ref{tab:t2i_speed}, VLDeformer achieves an average response time of $16$ms. $95$\% queries are finished real-time (less than $25$ms) and $99.99$\% within $100$ms. The response time of VinVL can be saved as batch size increases but with a sacrifice of space usage. In comparison, VLDeformer is stable in both the time and space cost because it enables encoding images offline so the main computations for one query are sentence encoding and cosine similarity operation.

\begin{table}
	\centering
	\small
	\setlength{\tabcolsep}{2.6pt}
	\begin{tabular}{@{}lcrrrrr|c@{}}
		\toprule
		\textbf{Model} & \textbf{Batch size} & \multicolumn{1}{c}{\textbf{Avg}} & \multicolumn{1}{c}{\textbf{Min}} & \multicolumn{1}{c}{\textbf{Max}} & \multicolumn{1}{c}{\textbf{P95}} & \multicolumn{1}{c|}{\textbf{P99.99}} & \textbf{Mem} \\ \midrule
		\multirow{3}{*}{VinVL} & 250 & 758 & 632 & 16634 & 846 & 10567 & 3.66 \\
		& 500 & 375 & 338 & 7172 & 398 & 6130 & 5.73 \\
		& 1000 & 197 & 168 & 3903 & 228 & 3542 & 9.84 \\ \midrule
		\multirow{3}{*}{VLDeformer} & 250 & 18 & 16 & 360 & 17 & 31 & 1.59 \\
		& 500 & 16 & 15 & 369 & 17 & 51 & 1.60 \\
		& 1000 & 16 & 15 & 382 & 17 & 33 & 1.60 \\ \bottomrule
	\end{tabular}
	\caption{Single Query Response time (ms) and corresponding memory space (GB) of single text-to-image retrieval query on 10k images. }
	\label{tab:t2i_speed}
\end{table}

\begin{figure}
	\centering
	\includegraphics[width=\linewidth]{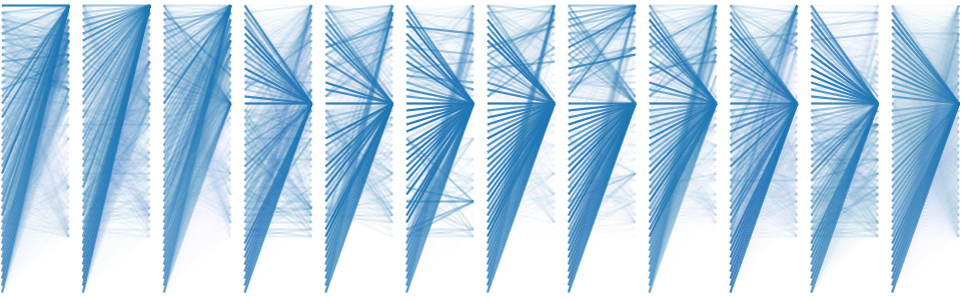}
	\includegraphics[width=\linewidth]{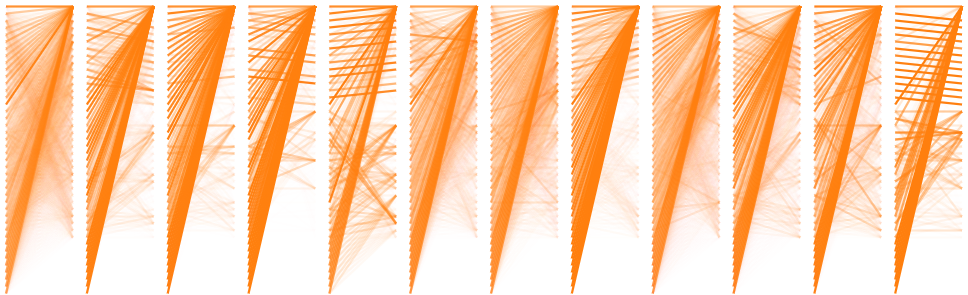}
	\includegraphics[width=\linewidth]{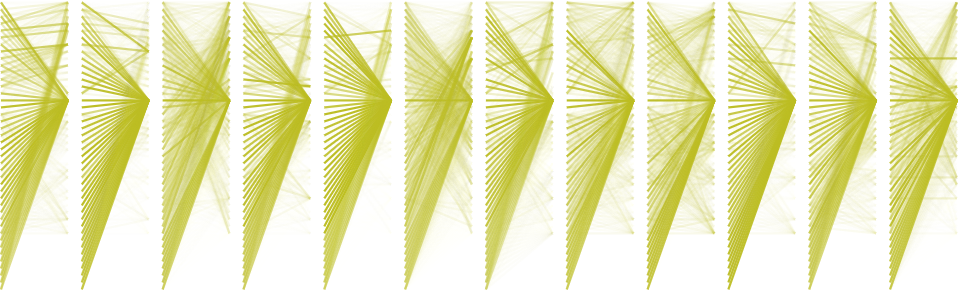}
	\caption{Illustration of the Uniter attention visualization. }
	\label{fig.uniter}
\end{figure}
\section{Detailed Qualitative Comparisons between VLDeformer and the backbone VinVL}
\begin{figure*}
	\centering
	\small
	\setlength{\tabcolsep}{0.1em}
	\begin{tabular}{ccccccccc}
		\includegraphics[width=2.4cm,height=2.4cm]{figures/case_study/t0.pdf} &
		\includegraphics[width=2.4cm,height=2.4cm]{figures/case_study/i0.png} & &
		\includegraphics[width=2.4cm,height=2.4cm]{figures/case_study/t0_i0.png} & 
		\includegraphics[width=2.4cm,height=2.4cm]{figures/case_study/t0_i1.png} &
		\includegraphics[width=2.4cm,height=2.4cm]{figures/case_study/t0_i2.png} &
		\includegraphics[width=2.4cm,height=2.4cm]{figures/case_study/t0_i3.png} & 
		\includegraphics[width=2.4cm,height=2.4cm]{figures/case_study/t0_i4.png} & \\
		\multicolumn{2}{c}{\textbf{VinVL result}:} & &
		\includegraphics[width=2.4cm,height=2.4cm]{figures/case_study/4096/1.png} &
		\includegraphics[width=2.4cm,height=2.4cm]{figures/case_study/4096/2.png} &
		\includegraphics[width=2.4cm,height=2.4cm]{figures/case_study/4096/3.png} & 
		\includegraphics[width=2.4cm,height=2.4cm]{figures/case_study/4096/4.png} & 
		\includegraphics[width=2.4cm,height=2.4cm]{figures/case_study/4096/5.png} &\\
		
		\includegraphics[width=2.4cm,height=2.4cm]{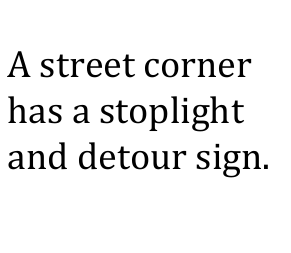} &
		\includegraphics[width=2.4cm,height=2.4cm]{figures/case_study/i1.png} & &
		\includegraphics[width=2.4cm,height=2.4cm]{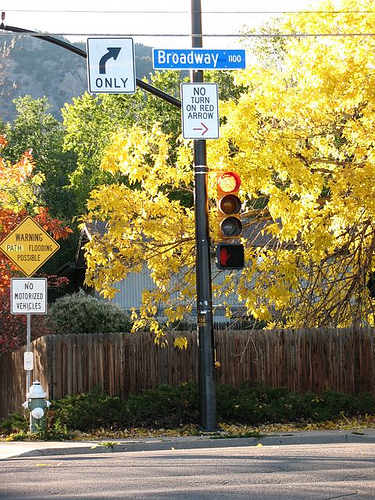} & 
		\includegraphics[width=2.4cm,height=2.4cm]{figures/case_study/t1_i1.png} &
		\includegraphics[width=2.4cm,height=2.4cm]{figures/case_study/t1_i2.png} &
		\includegraphics[width=2.4cm,height=2.4cm]{figures/case_study/t1_i3.png} & 
		\includegraphics[width=2.4cm,height=2.4cm]{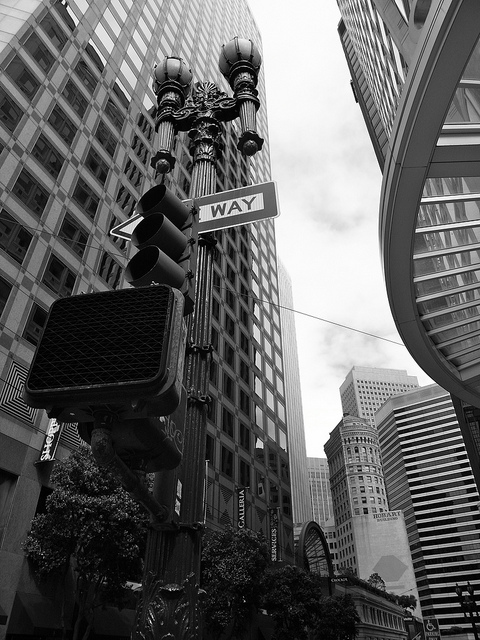} & \\
		\multicolumn{2}{c}{\textbf{VinVL result}:} & &
		\includegraphics[width=2.4cm,height=2.4cm]{figures/case_study/2/1.png} &
		\includegraphics[width=2.4cm,height=2.4cm]{figures/case_study/2/2.png} &
		\includegraphics[width=2.4cm,height=2.4cm]{figures/case_study/2/3.png} & 
		\includegraphics[width=2.4cm,height=2.4cm]{figures/case_study/2/4.png} & 
		\includegraphics[width=2.4cm,height=2.4cm]{figures/case_study/2/5.png} &\\
		
		\includegraphics[width=2.4cm,height=2.4cm]{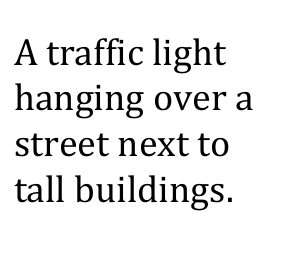} &
		\includegraphics[width=2.4cm,height=2.4cm]{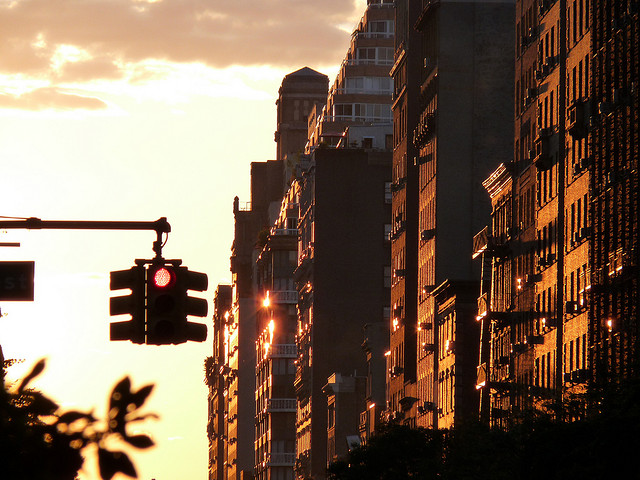} & &
		\includegraphics[width=2.4cm,height=2.4cm]{figures/case_study/t2_i0.png} & 
		\includegraphics[width=2.4cm,height=2.4cm]{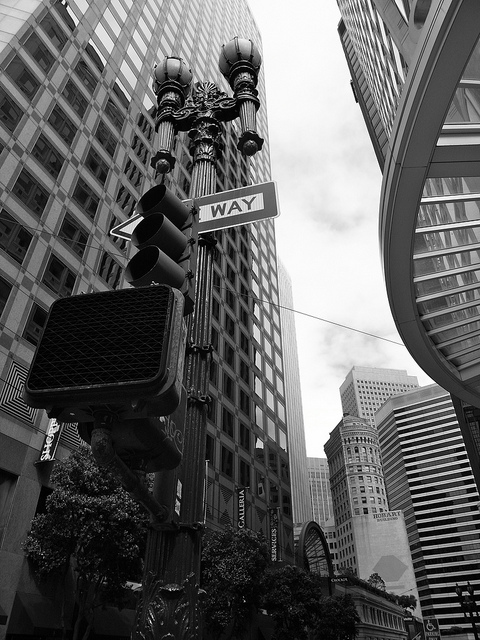} &
		\includegraphics[width=2.4cm,height=2.4cm]{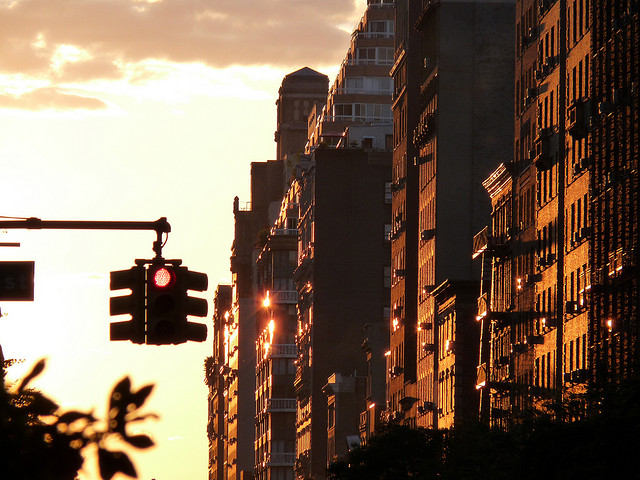} &
		\includegraphics[width=2.4cm,height=2.4cm]{figures/case_study/t2_i3.png} & 
		\includegraphics[width=2.4cm,height=2.4cm]{figures/case_study/t2_i4.png} & \\
		\multicolumn{2}{c}{\textbf{VinVL result}:} & &
		\includegraphics[width=2.4cm,height=2.4cm]{figures/case_study/4100/1.png} &
		\includegraphics[width=2.4cm,height=2.4cm]{figures/case_study/4100/2.png} &
		\includegraphics[width=2.4cm,height=2.4cm]{figures/case_study/4100/3.png} & 
		\includegraphics[width=2.4cm,height=2.4cm]{figures/case_study/4100/4.png} & 
		\includegraphics[width=2.4cm,height=2.4cm]{figures/case_study/4100/5.png} &\\
		
		\includegraphics[width=2.4cm,height=2.4cm]{figures/case_study/t3.pdf} &
		\includegraphics[width=2.4cm,height=2.4cm]{figures/case_study/i3.png} & & 
		\includegraphics[width=2.4cm,height=2.4cm]{figures/case_study/t3_i0.png} & 
		\includegraphics[width=2.4cm,height=2.4cm]{figures/case_study/t3_i1.png} &
		\includegraphics[width=2.4cm,height=2.4cm]{figures/case_study/t3_i2.png} &
		\includegraphics[width=2.4cm,height=2.4cm]{figures/case_study/t3_i3.png} & 
		\includegraphics[width=2.4cm,height=2.4cm]{figures/case_study/t3_i4.png} & \\
		\multicolumn{2}{c}{\textbf{VinVL result}:} & &
		\includegraphics[width=2.4cm,height=2.4cm]{figures/case_study/12/1.png} &
		\includegraphics[width=2.4cm,height=2.4cm]{figures/case_study/12/2.png} &
		\includegraphics[width=2.4cm,height=2.4cm]{figures/case_study/12/3.png} & 
		\includegraphics[width=2.4cm,height=2.4cm]{figures/case_study/12/4.png} & 
		\includegraphics[width=2.4cm,height=2.4cm]{figures/case_study/12/5.png} &\\
	\end{tabular}
	\caption{Top5 retrieved images from VLDeformer and VinVL that VLDeformer flips VinVL from right to wrong in R@$1$.  (Better viewed with zoom-in) }
	\label{fig.case1}
\end{figure*}

\begin{figure*}
	\centering
	\small
	\setlength{\tabcolsep}{0.1em}
	\begin{tabular}{ccccccccc}		
		\includegraphics[width=2.4cm,height=2.4cm]{figures/case_study/t6.pdf} &
		\includegraphics[width=2.4cm,height=2.4cm]{figures/case_study/i6.png} & & 
		\includegraphics[width=2.4cm,height=2.4cm]{figures/case_study/t6_i0.png} & 
		\includegraphics[width=2.4cm,height=2.4cm]{figures/case_study/t6_i1.png} &
		\includegraphics[width=2.4cm,height=2.4cm]{figures/case_study/t6_i2.png} &
		\includegraphics[width=2.4cm,height=2.4cm]{figures/case_study/t6_i3.png} & 
		\includegraphics[width=2.4cm,height=2.4cm]{figures/case_study/t6_i4.png} & \\
		\multicolumn{2}{c}{\textbf{VinVL result}:} & &
		\includegraphics[width=2.4cm,height=2.4cm]{figures/case_study/3801/1.png} &
		\includegraphics[width=2.4cm,height=2.4cm]{figures/case_study/3801/2.png} &
		\includegraphics[width=2.4cm,height=2.4cm]{figures/case_study/3801/3.png} & 
		\includegraphics[width=2.4cm,height=2.4cm]{figures/case_study/3801/4.png} & 
		\includegraphics[width=2.4cm,height=2.4cm]{figures/case_study/3801/5.png} &\\
		
		\includegraphics[width=2.4cm,height=2.4cm]{figures/case_study/t5.pdf} &
		\includegraphics[width=2.4cm,height=2.4cm]{figures/case_study/i5.png} & & 
		\includegraphics[width=2.4cm,height=2.4cm]{figures/case_study/t5_i0.png} & 
		\includegraphics[width=2.4cm,height=2.4cm]{figures/case_study/t5_i1.png} &
		\includegraphics[width=2.4cm,height=2.4cm]{figures/case_study/t5_i2.png} &
		\includegraphics[width=2.4cm,height=2.4cm]{figures/case_study/t5_i3.png} & 
		\includegraphics[width=2.4cm,height=2.4cm]{figures/case_study/t5_i4.png} & \\
		
		\multicolumn{2}{c}{\textbf{VinVL result}:} & &
		\includegraphics[width=2.4cm,height=2.4cm]{figures/case_study/4141/1.png} &
		\includegraphics[width=2.4cm,height=2.4cm]{figures/case_study/4141/2.png} &
		\includegraphics[width=2.4cm,height=2.4cm]{figures/case_study/4141/3.png} & 
		\includegraphics[width=2.4cm,height=2.4cm]{figures/case_study/4141/4.png} & 
		\includegraphics[width=2.4cm,height=2.4cm]{figures/case_study/4141/5.png} &\\
		
		\includegraphics[width=2.4cm,height=2.4cm]{figures/case_study/t4.pdf} &
		\includegraphics[width=2.4cm,height=2.4cm]{figures/case_study/i4.png} & & 
		\includegraphics[width=2.4cm,height=2.4cm]{figures/case_study/t4_i0.png} & 
		\includegraphics[width=2.4cm,height=2.4cm]{figures/case_study/t4_i1.png} &
		\includegraphics[width=2.4cm,height=2.4cm]{figures/case_study/t4_i2.png} &
		\includegraphics[width=2.4cm,height=2.4cm]{figures/case_study/t4_i3.png} & 
		\includegraphics[width=2.4cm,height=2.4cm]{figures/case_study/t4_i4.png} & \\
		
		\multicolumn{2}{c}{\textbf{VinVL result}:} & &
		\includegraphics[width=2.4cm,height=2.4cm]{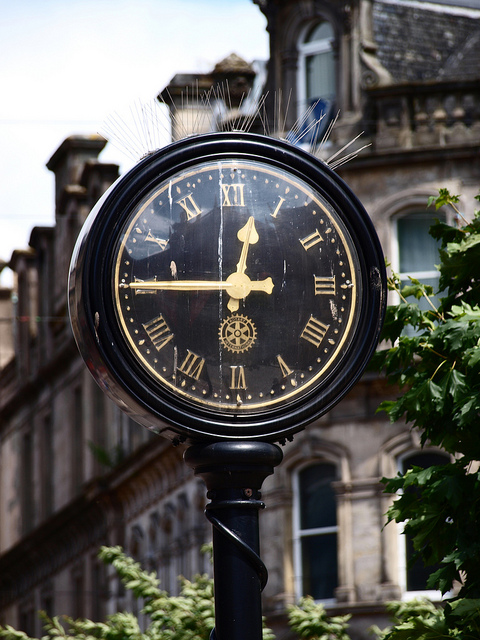} &
		\includegraphics[width=2.4cm,height=2.4cm]{figures/case_study/t4-o2.png} &
		\includegraphics[width=2.4cm,height=2.4cm]{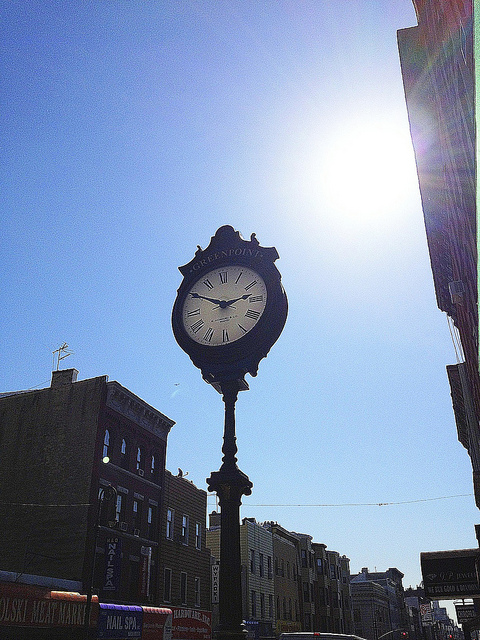} & 
		\includegraphics[width=2.4cm,height=2.4cm]{figures/case_study/t4-o4.png} & 
		\includegraphics[width=2.4cm,height=2.4cm]{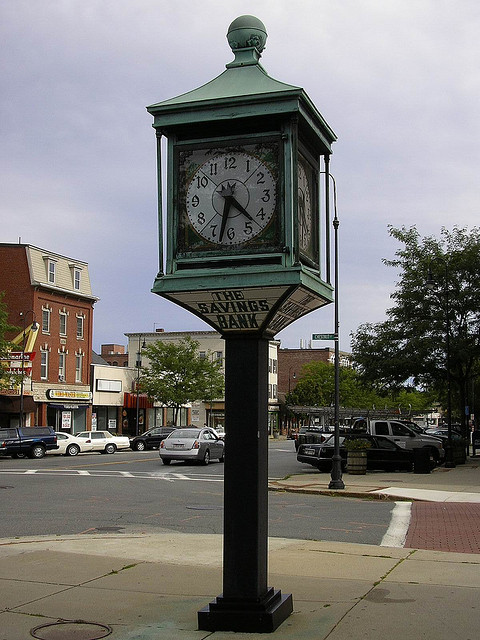} &\\
		Query Text &  Ground Truth  & & Top@1 & Top@2 & Top@3 & Top@4 & Top@5 
	\end{tabular}
	\caption{Top5 retrieved images from VLDeformer and VinVL that VLDeformer flips VinVL from right to wrong in R@$1$. For each comparison, the first row are VLDeformer results.(Better viewed with zoom-in) }
	\label{fig.case2}
\end{figure*}

Fig.\ref{fig.case1} and Fig.\ref{fig.case2} present more results that VLDeformer flipped the right results of VinVL to wrong. The top$5$ retrieved results of both VLDeformer and VinVL are presented. It can be found that the records of VLDeformer and VinVL are highly overlapped. Besides, although VLDeformer has a performance gap to the VinVL, both of the two models recall semantically similar images, indicating that the performance difference would decrease if we do not rely on the aligned image as the ground truth but using human evaluation that qualifies the real-world application requirements. More cases of VLDeformer are included in the Fig.\ref{fig.case3}. 

\section{Attention visualization cases}
In this section we show the attention visualization of Uniter. As the case shown in Figure.~\ref{fig.uniter}, there are two routing nodes in the Uniter attention dataflow. Similar to VinVL, the special tokens $\mathtt{[CLS]}$ and $\mathtt{[SEP]}$ are still routing nodes. We provide a notebook to view more cases online (See \textit{attention\_visualization\_notebook/} in the supplementary files). For both VinVL and uniter, we cached attention maps for $20$ samples.

\begin{figure*}
	\centering
	\setlength{\tabcolsep}{0.1em}
	\begin{tabular}{cccccccc}		
		\multicolumn{8}{l}{A man wearing a clown wig while riding on skis.} \\
		\includegraphics[width=2.5cm,height=2.5cm]{figures/appendix/1.png} & & 
		\includegraphics[width=2.5cm,height=2.5cm]{figures/appendix/10.png} & 
		\includegraphics[width=2.5cm,height=2.5cm]{figures/appendix/11.png} &
		\includegraphics[width=2.5cm,height=2.5cm]{figures/appendix/12.png} &
		\includegraphics[width=2.5cm,height=2.5cm]{figures/appendix/13.png} & 
		\includegraphics[width=2.5cm,height=2.5cm]{figures/appendix/14.png} & \\

		\multicolumn{8}{l}{a bench near a tree near a light pole.} \\
		\includegraphics[width=2.5cm,height=2.5cm]{figures/appendix/2.png} & & 
		\includegraphics[width=2.5cm,height=2.5cm]{figures/appendix/20.png} & 
		\includegraphics[width=2.5cm,height=2.5cm]{figures/appendix/21.png} &
		\includegraphics[width=2.5cm,height=2.5cm]{figures/appendix/22.png} &
		\includegraphics[width=2.5cm,height=2.5cm]{figures/appendix/23.png} & 
		\includegraphics[width=2.5cm,height=2.5cm]{figures/appendix/24.png} & \\

		\multicolumn{8}{l}{A delicious pizza sitting on a table next to a bottle of alcohol.} \\
		\includegraphics[width=2.5cm,height=2.5cm]{figures/appendix/3.png} & & 
		\includegraphics[width=2.5cm,height=2.5cm]{figures/appendix/30.png} & 
		\includegraphics[width=2.5cm,height=2.5cm]{figures/appendix/31.png} &
		\includegraphics[width=2.5cm,height=2.5cm]{figures/appendix/32.png} &
		\includegraphics[width=2.5cm,height=2.5cm]{figures/appendix/33.png} & 
		\includegraphics[width=2.5cm,height=2.5cm]{figures/appendix/34.png} & \\

		\multicolumn{8}{l}{A traffic light hanging over a street next to tall buildings.} \\
		\includegraphics[width=2.5cm,height=2.5cm]{figures/appendix/4.png} & & 
		\includegraphics[width=2.5cm,height=2.5cm]{figures/appendix/40.png} & 
		\includegraphics[width=2.5cm,height=2.5cm]{figures/appendix/41.png} &
		\includegraphics[width=2.5cm,height=2.5cm]{figures/appendix/42.png} &
		\includegraphics[width=2.5cm,height=2.5cm]{figures/appendix/43.png} & 
		\includegraphics[width=2.5cm,height=2.5cm]{figures/appendix/44.png} & \\

		\multicolumn{8}{l}{a man in a tie and a suit is indifferent.} \\
		\includegraphics[width=2.5cm,height=2.5cm]{figures/appendix/5.png} & & 
		\includegraphics[width=2.5cm,height=2.5cm]{figures/appendix/50.png} & 
		\includegraphics[width=2.5cm,height=2.5cm]{figures/appendix/51.png} &
		\includegraphics[width=2.5cm,height=2.5cm]{figures/appendix/52.png} &
		\includegraphics[width=2.5cm,height=2.5cm]{figures/appendix/53.png} & 
		\includegraphics[width=2.5cm,height=2.5cm]{figures/appendix/54.png} & \\

		\multicolumn{8}{l}{A skateboarder is taking a break at a skatepark..} \\
		\includegraphics[width=2.5cm,height=2.5cm]{figures/appendix/6.png} & & 
		\includegraphics[width=2.5cm,height=2.5cm]{figures/appendix/60.png} & 
		\includegraphics[width=2.5cm,height=2.5cm]{figures/appendix/61.png} &
		\includegraphics[width=2.5cm,height=2.5cm]{figures/appendix/62.png} &
		\includegraphics[width=2.5cm,height=2.5cm]{figures/appendix/63.png} & 
		\includegraphics[width=2.5cm,height=2.5cm]{figures/appendix/64.png} & \\

		\multicolumn{8}{l}{A bright, clean, modern kitchen sports track lighting and white cabinetry.} \\
		\includegraphics[width=2.5cm,height=2.5cm]{figures/appendix/7.png} & & 
		\includegraphics[width=2.5cm,height=2.5cm]{figures/appendix/70.png} & 
		\includegraphics[width=2.5cm,height=2.5cm]{figures/appendix/71.png} &
		\includegraphics[width=2.5cm,height=2.5cm]{figures/appendix/72.png} &
		\includegraphics[width=2.5cm,height=2.5cm]{figures/appendix/73.png} & 
		\includegraphics[width=2.5cm,height=2.5cm]{figures/appendix/74.png} & \\
		
		Ground Truth  & & Top@1 & Top@2 & Top@3 & Top@4 & Top@5 
	\end{tabular}
	\caption{More Top5 retrieved results where VLDeformer flips VinVL from right to wrong in R@$1$.  (Better viewed with zoom-in) }
	\label{fig.case3}
\end{figure*}